\documentclass{article} 
\usepackage[table,dvipsnames]{xcolor}
\usepackage[pdftex]{graphicx}
\usepackage{colm2024_conference}

\usepackage{microtype}
\usepackage{hyperref}
\usepackage{url}
\usepackage{booktabs}

\usepackage[some]{background}
\usepackage[algo2e]{algorithm2e} 

\SetBgScale{1}
\SetBgContents{\parbox{18cm}{%
  \Huge Potential Harmful Example\\[17cm]\rotatebox{180}{\Huge  }}}
\SetBgColor{red}
\SetBgAngle{270}
\SetBgOpacity{0.2}

\usepackage{url}

\usepackage{multirow}
\usepackage{subcaption}
\usepackage{enumitem}
\usepackage{wrapfig}
\usepackage{booktabs}       
\usepackage{tcolorbox}
\usepackage{amsmath}

\usepackage{array}
\newcommand{\PreserveBackslash}[1]{\let\temp=\\#1\let\\=\temp}
\newcolumntype{C}[1]{>{\PreserveBackslash\centering}p{#1}}

\usepackage{amssymb}
\usepackage{pifont}
\definecolor{hotpink}{RGB}{239,70,110}
\definecolor{darkyellow}{RGB}{191,144,0}

\colmfinalcopy

\title{HallE-Control: Controlling Object Hallucination in \\ Large Multimodal Models}

\author{Bohan Zhai$^1$\thanks{Equal Contribution} \And Shijia Yang$^{2*}$ \And Chenfeng Xu$^3$ \And Sheng Shen$^3$ \AND Kurt Keutzer$^3$  \And Chunyuan Li$^1$ \And Manling Li$^4$ \And\\
\texttt{ByteDance Inc.$^1$, Stanford University$^2$, UC Berkeley$^3$, UIUC$^4$} \\
}
\begin{document}

\maketitle

\begin{abstract}
Current Large Multimodal Models (LMMs) achieve remarkable progress, yet there remains significant uncertainty regarding their ability to accurately apprehend visual details, that is, in performing detailed captioning. To address this, we introduce \textit{CCEval}, a GPT-4 assisted evaluation method for detailed captioning. Interestingly, while LMMs demonstrate minimal object existence hallucination in existing VQA benchmarks, our proposed evaluation reveals continued susceptibility to such hallucinations. In this paper, we make the first attempt to investigate such hallucination from different aspects, including image resolution, the language decoder size, and instruction data amount, quality, granularity. Our findings underscore the unwarranted inference when the language description includes details at a finer object granularity than what the vision module can ground or verify, thus inducing hallucination. 
To control such hallucinations, we further attribute the reliability of captioning to contextual knowledge (involving only contextually grounded objects) and parametric knowledge (containing inferred objects by the model). 
Thus, we introduce $\textit{HallE-Control}$, a controllable LMM in terms of $\textbf{Hall}$ucination in object $\textbf{E}$xistence. HallE-Control can condition the captioning to shift between (i) exclusively depicting contextual knowledge for grounded objects and (ii) blending it with parametric knowledge to imagine inferred objects.
Our method reduces hallucination by 44\% compared to LLaVA$_{7B}$ and maintains the object coverage. Our code is publicly available at \url{https://github.com/bronyayang/HallE_Control}
\end{abstract}
\section{Introduction}
\vspace{-0.4cm}
\begin{figure*}[h]
  \centering
  \includegraphics[width=\linewidth]{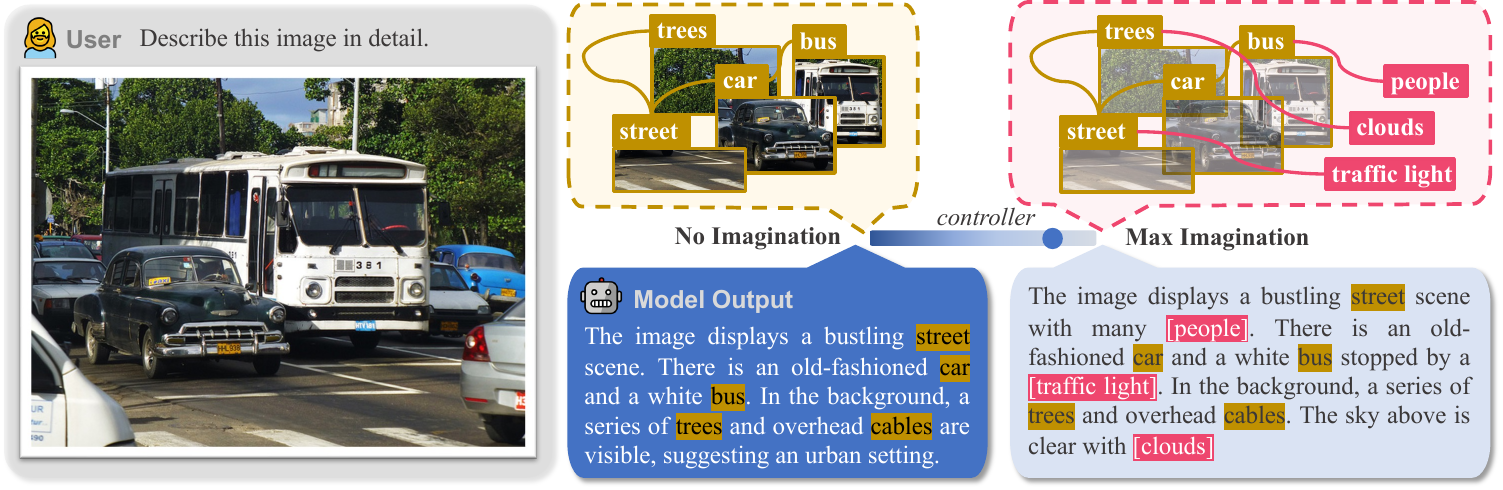}
  \captionsetup{aboveskip=5pt}\captionsetup{belowskip=0pt}\caption{HallE-Control uses a single continuous parameter during inference to control imagination in the caption. A control value of ``$-1$'' makes the model use solely contextual knowledge \colorbox{darkyellow}{(visually grounded objects)}, such as trees, buses, cars, and streets, whereas ``$+1$''makes the model incorporate parametric knowledge \colorbox{hotpink}{(inferred objects)}, such as people, clouds, and traffic lights, with the \texttt{[object]} marker labeling those inferred objects.}
  \label{fig:intro}
\end{figure*}

In recent years, Large Multimodal Models (LMMs)~\citep{liu2023visual, instructblip, li2023otter, zhu2023minigpt} have achieved significant progress, advancing tasks such as detailed captioning, visual conversations, and vision question-answering (VQA)~\citep{balanced_vqa_v2, MMBench, hudson2018gqa, fu2023mme}. However, similar to Large Language Models (LLMs)~\citep{touvron2023llama, MosaicML2023Introducing, openaichatgptblog} in the NLP domain, LMMs confront the issue of hallucination. This is particularly severe in detailed image captioning, which hinders the performance of downstream applications in robotics~\citep{huang2023voxposer}, visual search~\citep{hu2023avis}, etc.  To better understand and address this challenge, we first outline three types of object hallucinations frequently observed in the detailed captions:
(1) Object Existence Hallucination - The detailed image description references objects that are not present;
(2) Object Attribute Hallucination - The detailed image description inaccurately characterizes objects, misrepresenting attributes such as color, shape, and size;
(3) Object Relationship Hallucination - The detailed image description inaccurately depicts the relationships or interactions among objects, including erroneous relative positions, interaction states, and actions involving two or more objects. In this work, we mainly focus on \textit{defining the metric}, \textit{analyzing the cause}, and \textit{addressing the the problem of \textbf{object existence hallucination}}.

Evaluating detailed captions is inherently complex. Some of the efforts, including benchmarks like POPE~\citep{Li-hallucination-2023}, evaluate object hallucination using VQA. Such a bias towards VQA-based evaluations might result in an incomplete assessment of detailed captions, which requires obtaining a comprehensive view of visual details. To bridge this gap, we introduce \textit{CCEval}, designed specifically for object existence hallucination in detailed captions. To avoid the model gaining an unfair advantage by favoring shorter descriptions, CCEval maintains consistency in metrics such as average sentence length and the number of objects (see Sec. \ref{sec:analysis}). Notably, even models that well-performed on VQA-based object hallucination benchmarks showed substantial hallucinations when evaluated with CCEval.

In our exploration to uncover the underlying cause of object existence hallucination, we look into various factors including the size of the language decoder, the quantity, quality, and granularity of instruction data, and the input resolution to the vision encoder. We conclude the most crucial factor to be the alignment between objects mentioned in training caption and those vision encoder can perceive. During training of LMMs, the goal is to establish a one-to-one correspondence between objects mentioned in the caption and those present in the image. Objects successfully grounded by the vision encoder form accurate associations, internalizing them as \textbf{contextual knowledge}. Conversely, objects in the language that the vision encoder fails to ground create word-word semantic associations, which can be attributed to the generalization from the \textbf{parametric knowledge} within the model's parameters. During inference, when the model draws from such parametric knowledge, any misalignment can manifest as hallucination, as the model attempts to ``guess'' details not grounded by the vision module.

To address such hallucination, we are motivated by that \textit{not all hallucination is bad, and it is more desirable to control the generalization rather than outright removal of all imagined objects.} Recognizing the significance of both contextual and parametric knowledge in ensuring generation reliability, we present \textit{HallE-Control}, a novel approach to control the extent of expressed hallucination or parametric knowledge. We curate a 33k dataset similar to LLaVA~\citep{liu2023visual}, incorporating both pure contextual knowledge and a blend of contextual knowledge with marked parametric knowledge. Leveraging this dataset, we train a lightweighted single linear layer to control over the frozen LMM. As demonstrated in Figure \ref{fig:intro}, a singular continuous parameter adjustment (e.g. $-1 \rightarrow +1$) during inference enables the model to produce detailed captions with only contextual knowledge (e.g., $-1$) or blend with parametric knowledge (e.g., $+1$). Furthermore, the inferred objects from parametric knowledge are automatically highlighted with distinct tokens (e.g., \texttt{[object]}) for human reference. This method offers the advantage of preserving object count and coverage as well as sentence length, while effectively control object existence hallucination.

Overall, our contributions are:
\vspace{-0.3cm}
\begin{itemize}
\item A comprehensive analysis on LMM components that influence hallucination, with a specific focus on alignment issues in the vision encoder and instruction data;
\item The first approach to control object existence hallucination within detailed captions;
\item A novel evaluation method for detailed caption object existence hallucination, with metrics such as object count, coverage, and average sentence length, alongside hallucination assessment.
\end{itemize}

\section{Hallucination Analysis}
\label{sec:analysis}
Object existence hallucination can be influenced by several factors, including the language decoder, instruction data, and vision encoder. In our analysis, we address each factor individually. For a diverse methodological exploration, we select LLaVA, InstructBLIP~\citep{instructblip}, and Shikra~\citep{chen2023shikra}: LLaVA and Shikra share the same model structure; Shikra and InstructBLIP use mixed-dataset and multi-task instruction data; InstructBLIP finetunes only Q-former, while the other finetune projector and LLM. More details about models can be found in Appendix.

\subsection{Benchmarks}
There are two primary methods, VQA-based and caption-based benchmarks, for evaluating object existence hallucination in LMMs.

\begin{table*}
\centering
\captionsetup{aboveskip=0pt}\captionsetup{belowskip=0pt}
\caption{We evaluate LLaVA Vicuna$_{7B}$, LLaVA Vicuna$_{13B}$, Shikra$_{7B}$, InstructBLIP Vicuna$_{7B}$ public checkpoints on VQA-based benchmarks, including POPE and MME.}
\label{tab:vqa}
\tiny

\resizebox{10cm}{!}{
\begin{tabular}
{l|l|c@{\hspace{1.0\tabcolsep}}c@{\hspace{1.0\tabcolsep}}c@{\hspace{1.0\tabcolsep}}c@{\hspace{1.0\tabcolsep}}c@{\hspace{1.0\tabcolsep}}}
\toprule
\textbf{Benchmark} & \textbf{Model} & \textbf{Accuracy}$\uparrow$ & \textbf{Precision}$\uparrow$ & \textbf{Recall}$\uparrow$ & \textbf{F1}$\uparrow$ & \textbf{Yes (\%)}\\
\midrule
\multirow{4}{*}{POPE - Random} & LLaVA$_{\textsc{7B}}$ & 73.13 & 66.95 & 94.53 & 78.39 & 72.78 \\
& LLaVA$_{\textsc{13B}}$ & 78.49 & 73.57 & 90.93 & 81.34 & 63.71 \\
& Shikra$_{\textsc{7B}}$ & 86.99 &	94.77 &	79.13 &	86.25 &	43.04 \\
& InstructBLIP$_{\textsc{7B}}$ & 86.60 & 80.74	& 96.13	& 87.77	& 59.53\\
\midrule
\multirow{4}{*}{POPE - Popular} & LLaVA$_{\textsc{7B}}$ & 59.87 & 55.88 & 93.80 & 70.03 & 83.93 \\
& LLaVA$_{\textsc{13B}}$ & 70.80 & 64.73 &	91.40 &	75.79 &	70.60 \\
& Shikra$_{\textsc{7B}}$ & 84.35 & 88.10 &	79.43 &	83.54 &	45.08 \\
& InstructBLIP$_{\textsc{7B}}$ & 71.27 & 64.20 & 96.13	& 76.99 & 74.87 \\
\midrule
\multirow{4}{*}{POPE - Adversarial} & LLaVA$_{\textsc{7B}}$ & 57.06 & 54.07 & 93.93 & 68.63 & 86.87 \\
& LLaVA$_{\textsc{13B}}$ & 63.93 &	59.03 &	91.07 & 71.63 & 77.13 \\
& Shikra$_{\textsc{7B}}$ & 82.88 & 85.32 &	79.43 & 82.27 &	46.55 \\
& InstructBLIP$_{\textsc{7B}}$ & 72.10	& 65.13	& 95.13 & 77.32 & 73.03 \\
\toprule
\textbf{Benchmark} & \textbf{Model} & \textbf{Existence}$\uparrow$ & \textbf{Count}$\uparrow$ & \textbf{Position}$\uparrow$ & \textbf{Color}$\uparrow$ & \textbf{Total}$\uparrow$\\
\midrule
\multirow{4}{*}{MME} & LLaVA$_{\textsc{7B}}$ & 150.00 & 48.33 & 50.00 & 55.00 & 303.33 \\
& LLaVA$_{\textsc{13B}}$ & 180.00 & 113.33 & 55.00 & 95.00 & 443.33 \\
& Shikra$_{\textsc{7B}}$ & 185.00 & 118.33 & 75.00 & 155.00 & 533.33 \\
& InstructBLIP$_{\textsc{7B}}$ & 185.00 & 60.00 & 50.00 & 125.00 & 420.00 \\

\bottomrule
\end{tabular}
}
\end{table*}
\textbf{VQA-based benchmarks} pose questions about objects within images. For a model to be considered hallucination-free, it should address these visual questions accurately. Notably, a large proportion of questions are simply binary, typically asking about the presence or attributes of objects.

The POPE benchmark evaluates object existence hallucination by a polling-based query method, consisting of a series of yes/no questions on sampled objects from visual instructions. POPE contains three sets: random, popular, and adversarial. These subsets respectively focus on randomly selected objects, frequently occurring objects, and those objects that co-occur in training sets. We choose POPE evaluation on MSCOCO~\citep{lin2014microsoft} dataset. MME~\citep{fu2023mme} coarse-grained recognition construct yes/no questions similarly but selects objects at random. This benchmark has 30 images, with each image paired with two questions: one positive and one negative.

In Table \ref{tab:vqa}, LLaVA${_{7B}}$ exhibits the greatest degree of hallucination, whereas Shikra outperforms other models in both POPE and MME. Specifically, Shikra shows a significantly higher F1 score in both POPE-popular and POPE-adversarial categories, while LLaVA${_{7B}}$ displays the lowest. Additionally, Shikra's "Yes" ratio is closer to a balanced 50\% compared to other models. However, in subsequent sections, we demonstrate that these observations from VQA-based benchmarks are not consistent with those from caption-based benchmarks.

\begin{table*}
\centering
\vspace{-1em}
\captionsetup{aboveskip=2pt}\captionsetup{belowskip=0pt}
\caption{Comparison between CHAIR and our evaluation method, CCEval.}
\label{tab:our_eval}

\resizebox{\textwidth}{!}{
\begin{tabular}
{l|c@{\hspace{1.0\tabcolsep}}c@{\hspace{1.0\tabcolsep}}c@{\hspace{1.0\tabcolsep}}c@{\hspace{1.0\tabcolsep}}|c@{\hspace{1.0\tabcolsep}}c@{\hspace{1.0\tabcolsep}}c@{\hspace{1.0\tabcolsep}}c@{\hspace{1.0\tabcolsep}}c@{\hspace{1.0\tabcolsep}}c@{\hspace{1.0\tabcolsep}}}
\toprule
\multirow{2}{*}{\textbf{Model}} & \multicolumn{4}{c@{\hspace{1.0\tabcolsep}}|}{\textbf{CHAIR}} & \multicolumn{5}{c@{\hspace{1.0\tabcolsep}}}{\textbf{CCEval (Ours)}} \\
& \textbf{\textit{CHAIR}$_s$}$\downarrow$ & \textbf{\textit{CHAIR}$_i$}$\downarrow$ & \textbf{Avg. Length}$\uparrow$ & \textbf{Avg. Object}$\uparrow$ & \textbf{\textit{CHAIR}$_s$}$\downarrow$ & \textbf{\textit{CHAIR}$_i$}$\downarrow$ & \textbf{Coverage}$\uparrow$ & \textbf{Avg. Length}$\uparrow$ & \textbf{Avg. Object}$\uparrow$\\
\midrule
LLaVA$_{7B}$ & 24.1 & 9.1 & 42.5 & 3.7 & 72.00 & 19.7 & 32.74 & 92.27 & 9.19 \\
LLaVA$_{13B}$ & 60.6 & 18.4 & 90.2 & 7.6 & 79.00 & 23.80 & 33.56 & 108.02 & 9.28 \\
Shikra$_{7B}$ & 59.1 & 16.6 & 91.2 & 7.5 & 83.00 & 24.40 & 33.29 &109.37 & 9.10 \\
InstructBLIP$_{7B}$ & 1.4 & 1.7 & 2.3 & 0.8 & 72.00 & 22.30 & 29.76 & 108.42 & 8.04 \\
\bottomrule
\end{tabular}
}
\end{table*}

\textbf{Caption-based benchmarks}, like CHAIR, begin by splitting the sentence and extracting nouns. Subsequently, it augments the ground truth objects by incorporating hard-coded synonyms and phrases, forming a ground truth set. The benchmark then identifies hallucinated objects by comparing the objects in the caption with this ground truth set. CHAIR computes $\textsc{CHAIR}_i$ and $\textsc{CHAIR}_s$ as follows:
$$\textsc{CHAIR}_i = \frac{|\{\text{hallucinated\ objects}\}|}{|\{\text{all\ objects\ mentioned}\}|}$$
$$\textsc{CHAIR}_s = \frac{|\{ \text{sentences\ with\ hallucinated\ object\}}|}{|\{\text{all\ sentences\}}|}$$

Table \ref{tab:our_eval}(left) reveals that while InstructBLIP exhibits minimal object existence hallucination, it averages a mere 0.8 objects per sentence. In contrast, LLaVA$_{13B}$ and Shikra manifest more hallucination, but they also generate more detailed captions, with as many as 7.6 and 7.5 objects per sentence, respectively. We find comparing object hallucinations is impractical when there is a significant disparity in average sentence length and number of objects.

Apart from these disparities, the use of a hard-coded ground truth set is another challenge. To counter these challenges, we introduce \textit{CCEval}, a GPT-4 assisted evaluation for detailed captions.  We first prompt LMMs to generate detailed captions on 100 randomly sampled images from Visual Genome~\citep{krishna2017visual}. Subsequently, utilizing GPT-4's in-context learning capabilities, we extract individual objects from these captions and identify hallucinated ones by referencing the provided ground truth objects. On top of CHAIR ~\cite{objectHallucination}, we introduce "coverage" metric to ensure that the captions are detailed enough. This metric computes the ratio of objects in the caption that match the ground truth to the total number of ground truth objects. We additionally record and balance the average number of objects as well as the average length of captions across all cases. More details of CCEval can be found in Appendix.

As reflected in Table \ref{tab:our_eval}(right), when subjected to consistent constraints—average sentence length approximately 100 words and around 9 objects per sentence—all models exhibit comparably sub-optimal results. Interestingly, while Shikra surpass other models in VQA-based benchmarks, especially in the POPE , it under-performs in CCEval. This suggests that hallucination in detailed captions is not consistently captured by VQA-based evaluations.

\begin{table*}
\centering
\captionsetup{aboveskip=2pt}\captionsetup{belowskip=0pt}
\caption{Performance of LLaVA and InstructBLIP with different sizes of language decoder. LLaVA are trained on CC-595k for stage one and Instruction-150k for stage two.}
\label{tab:lang}
\tiny

\resizebox{12cm}{!}{
\begin{tabular}
{l|l|c@{\hspace{1.0\tabcolsep}}c@{\hspace{1.0\tabcolsep}}c@{\hspace{1.0\tabcolsep}}c@{\hspace{1.0\tabcolsep}}c@{\hspace{1.0\tabcolsep}}}
\toprule
\textbf{Benchmark} & \textbf{Model} & \textbf{Accuracy}$\uparrow$ & \textbf{Precision}$\uparrow$ & \textbf{Recall}$\uparrow$ & \textbf{F1}$\uparrow$ & \textbf{Yes (\%)}\\
\midrule
\multirow{5}{*}{POPE - Random} & LLaVA$_\textsc{7B}$ & 75.77 & 69.79 &	93.47 & 79.91 & 69.04\\
& LLaVA$_\textsc{13B}$ & 78.49 & 73.57 & 90.93 & 81.34 & 63.71 \\
& LLaVA$_\textsc{33B}$ & 78.14 & 73.18 & 90.93 & 81.09 & 64.05\\
& InstructBLIP$_\textsc{7B}$ & 86.60 & 80.74 & 96.13 & 87.77 & 59.53 \\
& InstructBLIP$_\textsc{13B}$ & 88.73 & 86.67 & 92.33 & 89.41 & 54.91 \\
\midrule
\multirow{5}{*}{POPE - Popular} & LLaVA$_{7B}$ & 65.07 & 59.60 & 93.53 & 72.81 & 78.47 \\
& LLaVA$_\textsc{13B}$ & 70.80 & 64.73 & 91.40	& 75.79 & 70.60 \\
& LLaVA$_\textsc{33B}$ & 72.43 & 66.45 & 90.60 & 76.67 & 68.17 \\
& InstructBLIP$_\textsc{7B}$ & 71.27 & 64.20 & 96.13 & 76.99 & 74.87 \\
& InstructBLIP$_\textsc{13B}$ & 80.53 & 74.70 & 92.33 & 82.59 & 61.80 \\
\midrule
\multirow{5}{*}{POPE - Adversarial} & LLaVA$_{7B}$ & 57.07 & 54.07 & 93.93 & 68.63 & 86.87  \\
& LLaVA$_\textsc{13B}$ & 63.93 & 59.03 & 91.07 & 71.63 & 77.13 \\
& LLaVA$_\textsc{33B}$ & 66.30	& 60.91 & 91.00	& 72.98 & 74.70 \\
& InstructBLIP$_\textsc{7B}$ & 72.10 & 65.13 & 95.13 & 77.32 & 73.03 \\
& InstructBLIP$_\textsc{13B}$ & 73.97 & 67.53 & 92.33 & 78.01 & 68.37 \\
\toprule
\textbf{Benchmark} & \textbf{Model} & \textbf{\textit{CHAIR}$_s$}$\downarrow$ & \textbf{\textit{CHAIR}$_i$}$\downarrow$ & \textbf{Coverage}$\uparrow$ & \textbf{Avg. Length}$\uparrow$ & \textbf{Avg. Object}$\uparrow$\\
\midrule
\multirow{5}{*}{CCEval (Ours)} & LLaVA$_{7B}$ & 82.00 & 25.30 & 33.58 & 109.89	& 9.31 \\
& LLaVA$_\textsc{13B}$ & 79.00 & 23.80 & 33.56 & 108.02 & 9.28\\
& LLaVA$_\textsc{33B}$ & 82.00 & 21.80 & 31.26 & 106.85 & 9.07 \\
& InstructBLIP$_\textsc{7B}$ & 72.00 & 22.30 & 29.76 & 108.42 & 8.04\\
& InstructBLIP$_\textsc{13B}$ & 64.00 & 16.70 & 33.60 & 101.63 & 8.06\\

\bottomrule
\end{tabular}
}
\end{table*}

\subsection{Language Decoder}

We investigate if expanding the size of the language backbone can mitigate object existence hallucination. As detailed in Table \ref{tab:lang}, the language decoder of LLaVA is increased from 7B to 33B, and for InstructBLIP, it is increased from 7B to 13B. The result shows that hallucination for LLaVA reduced for POPE but not for CCEval. For InstructBLIP, $\textsc{CHAIR}_i$ and $\textsc{CHAIR}_s$ is reduced by 8 and 5.6 on CCEval, respectively. However, although there is a gain for scaling up language backbone, it is not consistent or salient from the observation, suggesting language decoder is not a primary factor in reducing hallucination.

\subsection{Data}
Similar to our approach with the language decoder, we begin by scaling up the volume of instruction finetuning data, ranging from 80K to 2.4M. As illustrated in Table \ref{tab: data}, the LLaVA$_{7B}$ model, finetuned on 80K instruction data, exhibits fewer object existence hallucinations compared to the models finetuned on 150K and SVIT~\citep{zhao2023svit}. The result suggests extra data without quality guarantee may increase hallucination for both VQA-based and caption-based evaluations. Given that all three datasets are generated by GPT-4, we question the quality of the data. LRV also raises this concern, suggesting that the training data itself might contain hallucinations. Some examples of training data are presented in the Appendix. Interestingly, we find that GPT-4 does not introduce additional object existence hallucination at all -- the issue lies in MSCOCO ground truth objects that are given to GPT-4 and to be asked to strictly include in generated captions. These GT objects are challenging even for human observers to ground, due to factors like size, resolution, and occlusion. This led us to hypothesize that the vision encoder might also struggle to ground these objects effectively.
\begin{table*}
\centering
\captionsetup{aboveskip=2pt}\captionsetup{belowskip=0pt}
\caption{Performance of LLaVA$_{7B}$ with different sizes of data. 80K and 158K contains 80K and 158K data respectively, and SVIT contains 2.4M.}
\label{tab: data}
\tiny

\resizebox{12cm}{!}{
\begin{tabular}
{l|l|c@{\hspace{1.0\tabcolsep}}c@{\hspace{1.0\tabcolsep}}c@{\hspace{1.0\tabcolsep}}c@{\hspace{1.0\tabcolsep}}c@{\hspace{1.0\tabcolsep}}}
\toprule
\textbf{Benchmark} & \textbf{Finetune Data} & \textbf{Accuracy}$\uparrow$ & \textbf{Precision}$\uparrow$ & \textbf{Recall}$\uparrow$ & \textbf{F1}$\uparrow$ & \textbf{Yes (\%)}\\
\midrule
\multirow{3}{*}{POPE - Random} & 80K & 73.13 & 66.95 & 94.53 & 78.39 & 72.78 \\
& 158K & 75.77 & 69.79 & 93.47 & 79.91 & 69.04 \\
& SVIT & 52.34 & 52.00 & 97.87 & 67.92 & 97.01 \\
\midrule
\multirow{3}{*}{POPE - Popular} & 80K &	59.87 & 55.88 & 93.80 & 70.03 & 83.93 \\
& 158K & 65.07 & 59.60 & 93.53 & 72.81 & 78.47 \\
& SVIT & 50.77 & 50.43 & 90.47 & 64.76 & 89.70 \\
\midrule
\multirow{3}{*}{POPE - Adversarial} & 80K & 57.07 & 54.07 & 93.93 & 68.63 & 86.87 \\
& 158K & 58.47 & 55.00 & 93.07 & 69.14 & 84.6 \\				
& SVIT & 51.37 & 50.77 & 90.33 & 65.00 & 88.97 \\					
\toprule
\textbf{Benchmark} & \textbf{Finetune Data} & \textbf{\textit{CHAIR}$_i$}$\downarrow$ & \textbf{\textit{CHAIR}$_s$}$\downarrow$ & \textbf{Coverage}$\uparrow$ & \textbf{Avg. Length}$\uparrow$ & \textbf{Avg. Object}$\uparrow$\\
\midrule
\multirow{3}{*}{CCEval (Ours)} & 80K & 72.00 & 19.70  & 32.74 & 92.27 & 9.19 \\
& 158K & 82.00 & 25.30 & 33.58 & 109.89 & 9.31 \\
& SVIT & 87.00 & 23.30 & 47.46 & 296.63 & 18.14 \\

\bottomrule
\end{tabular}
}
\end{table*}
\begin{table*}
\centering
\captionsetup{aboveskip=2pt}\captionsetup{belowskip=0pt}
\caption{Performance of LLaVA with Llama 2$_{13B}$ language decoder and CLIP-Large vision encoder with different input resolutions.}
\label{tab:resolution_13B}
\tiny

\resizebox{12cm}{!}{
\begin{tabular}
{l|l|c@{\hspace{1.0\tabcolsep}}c@{\hspace{1.0\tabcolsep}}c@{\hspace{1.0\tabcolsep}}c@{\hspace{1.0\tabcolsep}}c@{\hspace{1.0\tabcolsep}}}
\toprule
\textbf{Benchmark} & \textbf{Vision Encoder} & \textbf{\textit{CHAIR}$_s$}$\downarrow$ & \textbf{\textit{CHAIR}$_i$}$\downarrow$ & \textbf{Coverage}$\uparrow$ & \textbf{Avg. Length}$\uparrow$ & \textbf{Avg. Object}$\uparrow$\\
\midrule
\multirow{3}{*}{CCEval (Ours)} & CLIP-L-112x & 79.00 & 21.70 & 32.04 & 110.36 & 9.12 \\
& CLIP-L-224x & 74.00 & 19.30 & 32.83 & 113.03 & 9.18 \\
& CLIP-L-336x & 64.00 & 16.00 & 33.37 & 108.52 & 8.92 \\
\bottomrule
\end{tabular}
}
\end{table*}
\begin{table*}
\centering
\captionsetup{aboveskip=2pt}\captionsetup{belowskip=0pt}
\caption{Performance of LLaVA$_{7B}$ and with sliding window technique (SW).}
\label{tab:resolution_7B}
\tiny

\resizebox{12cm}{!}{
\begin{tabular}
{l|l|c@{\hspace{1.0\tabcolsep}}c@{\hspace{1.0\tabcolsep}}c@{\hspace{1.0\tabcolsep}}c@{\hspace{1.0\tabcolsep}}c@{\hspace{1.0\tabcolsep}}}
\toprule
\textbf{Benchmark} & \textbf{Vision Encoder} & \textbf{\textit{CHAIR}$_s$}$\downarrow$ & \textbf{\textit{CHAIR}$_i$}$\downarrow$ & \textbf{Coverage}$\uparrow$ & \textbf{Avg. Length}$\uparrow$ & \textbf{Avg. Object}$\uparrow$\\
\midrule
\multirow{3}{*}{CCEval (Ours)} & CLIP-L-224x & 79.00 & 21.70 & 32.04 & 110.36 & 9.12 \\
& CLIP-L-336x & 79.00 & 18.90 & 36.00 & 111.55 & 9.19 \\
& CLIP-L-224x (SW) & 72.00 & 18.70 & 36.89 & 110.43 & 8.65 \\
\bottomrule
\end{tabular}
}
\end{table*}
\subsection{Vision Encoder}

Intuitively, increasing image resolution enhances model’s perception of finer details, thus making the grounding of objects mentioned in the caption easier. To verify our hypothesis from the previous section, we increase the input image resolution for the vision encoder. Specifically, for our evaluation, the resolution for LLaVA$_{7B}$ was incremented from 224x to full resolution using a sliding window approach for efficiency, as detailed in the Appendix. Table \ref{tab:resolution_7B} shows a constant decrease in hallucination and increase in object coverage. Additionally, we assess LLaVA with Llama 2$_{13B}$, varying the resolution from 112x to 336x. For the 112x112 resolution, the original image was downscaled to 112x and subsequently upscaled to 224x before being input to CLIP-Large-224x. Table \ref{tab:resolution_13B} gives a consistent observation that larger input resolution can reduce hallucination in image captions.

 \textbf{Conclusion.} Through our systematic analysis of object existence hallucination, we summarize several insights: (1) Enlarging the language decoder does mitigate hallucination, but improvements are not huge.  (2) Expanding the volume of instruction data actually increases hallucination. Upon inspection of training data, we find certain objects described in the captions might not be grounded by the vision encoder. (3) To validate our hypothesis in (2), we show improving input image resolution significantly reduces hallucination by enhancing model grounding ability.

Reflecting on these findings, we attempt to provide an explanation for the reason of object existence hallucination in detailed captions. The process of image captioning in LMMs can be perceived as a form of information mapping or translation. Ideally, the goal is to have a direct one-to-one correspondence between objects identified in the image and those mentioned in the captions. Objects successfully grounded by the vision encoder form accurate correspondence, making this as \textbf{contextual knowledge} in the model, following~\citep{neeman-etal-2023-disentqa}. When objects in training caption fail to ground by the vision encoder, the model learns \textbf{parametric knowledge}, the knowledge encoded in the model's parameter. This kind of knowledge is the association of objects in the language with other words instead of with corresponding image object feature. During inference, when the model draws from parametric knowledge, it attempts to ``guess'' details not grounded by the vision module and is perceived as object existence hallucination. 

\section{Hallucination Controlling}
\begin{figure*}[h]
  \centering
  \vspace{-0em}
  \includegraphics[width=1.0\linewidth]{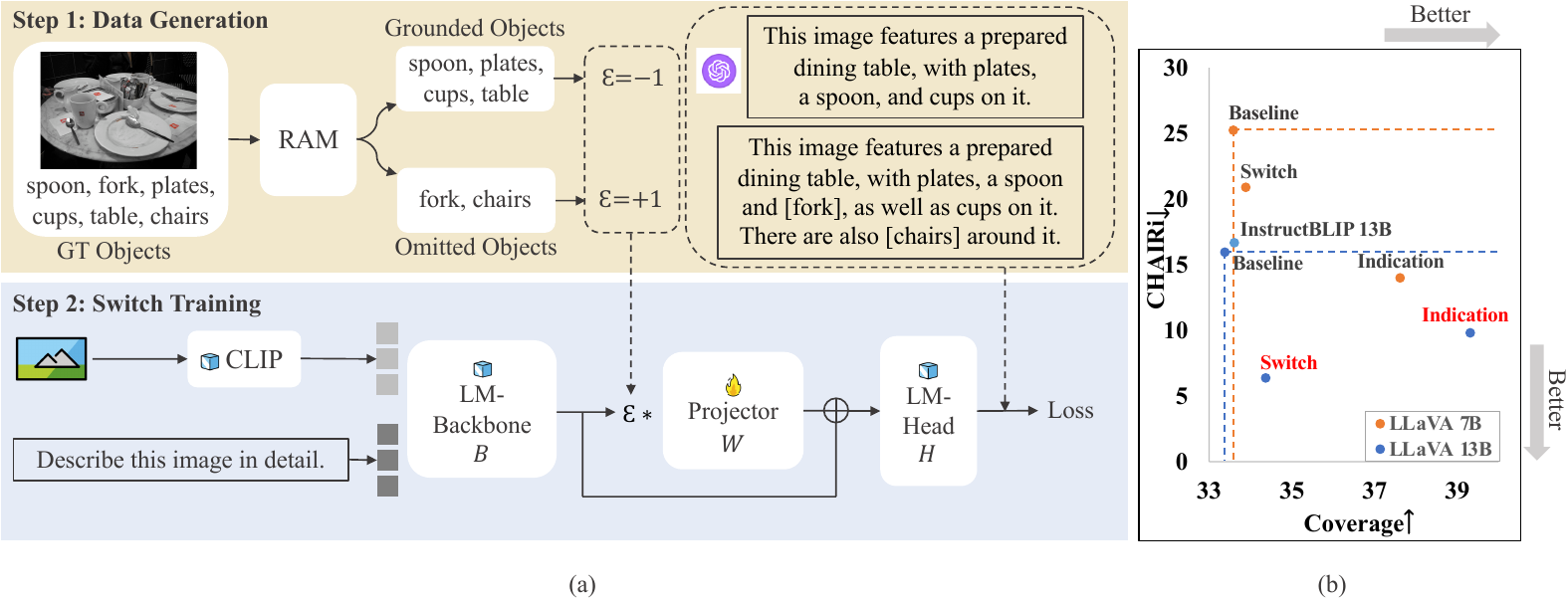}
  \captionsetup{aboveskip=5pt}\captionsetup{belowskip=0pt}\caption{ (a) shows the overall training pipeline of HallE-Control. When generating data, we use RAM to separate ground truth objects to visually grounded and omitted objects. Then, we utilize GPT-4 to convert this existing list of grounded objects into a caption as contextual data, and we assign $\varepsilon$ as $-1$. We put bracket around omitted objects in the original LLaVA caption as parametric joint data and assign $\varepsilon$ as $+1$. During training, we supervise using contextual only data and parametric joint data, pass in $\varepsilon$ as $-1$ or $+1$, respectively. (b) shows our methods consistently outperforms LLaVA baselines and InstructBLIP$_{13B}$. Indication is w/ ind. in Table \ref{tab:special_token}, and the Control is -1 in Table \ref{tab:switch}.}
  \label{fig:method}
\end{figure*}




Building on our prior finding that parametric knowledge can lead to hallucination, controlling or signaling the model's use of this knowledge enables us to manage its tendency for hallucination. By incorporating a controller layer, we directly influence the output distribution linked to parametric knowledge. To this end, we developed \textit{HallE-Control}, a LMM designed to control the extent of parametric knowledge within detailed captions. For this purpose, we developed two datasets to train the controller: the first captures solely contextual knowledge, while the second merges both contextual and parametric knowledge. Additionally, we can make the model signals the parametric knowledge by marking unrecognised objects in the finetuning dataset.

\subsection{Data Generation}
\label{sec:switch-dataset}
\textbf{Grouping using RAM.} We begin by passing MSCOCO's ground truth objects to the open vocabulary detector RAM~\citep{zhang2023recognize}. RAM categorizes the objects into two groups: ``grounded'' (detected objects as contextual group) and ``omitted'' (not detected objects as parametric group). This step aims to simulate the maximum visual granularity achievable by a vision encoder in LMMs.

\textbf{Contextual Data Generation.} Our first dataset involves generating detailed captions using only objects from the contextual group. To do this, we feed MSCOCO source labels (including object classes, bounding boxes, and short captions) into GPT-4. We adhere to LLaVA's caption creation pipeline and provide prompt in Appendix.

\textbf{Parametric Joint Data Generation.} The second dataset incorporates both contextual and parametric knowledge. Here, we begin with LLaVA's original detailed captions and annotate objects from the parametric group with special tokens. Specifically, we enclose the ``omitted'' objects with brackets. Formally, if $S$ denotes the original image caption sentence and $X = \{x_1, ... , x_n\}$ represents a set of undetected objects, our data processing can be represented as:
   $$
   S_{new} = \text{replace}(S, x_i, [x_i])
   $$
   The purpose of bracketing the parametric objects is twofold: it serves as an indicator during inference and provides a hint during training.

\subsection{Hallucination Control}

Inspired by LM-Switch~\citep{han2023lm}, we address hallucination by adding a control parameter $\varepsilon$, e.g., $+1$ for permitted imagination and $-1$ for restricting imagination, 
as depicted in Figure \ref{fig:method}(a). Let \( M \) represent the LLM with fixed parameters: $M(x)  = H(e_v)$ where  $H$ stands for LM-head and  $e_v=B(x)$ is the output word embedding from LM-backbone. We modify $M^{\prime}=M(\varepsilon W)$, thus making output word embedding $e_v^{\prime}= e_v + \varepsilon W e_v$, leading to the derived model $M^{\prime}$ as: 
\begin{equation*}
    M^{\prime}(x) = H(B(x) + \varepsilon W(B(x))).
\end{equation*}

 The learned projector $W$ can be regarded as the transformation from a generic word space to the object sensitive word space, where word-word semantic correspondence is optimized to object correspondence, and $\varepsilon$ governs the intensity of imagining related objects. 

\textbf{Training.} To train such a controlling parameter, we leverage the contrastive training data covering both contextual and parametric datasets in Sec~\ref{sec:switch-dataset}. For data with only contextual knowledge, we assign \( \varepsilon = -1 \) when inputting into the model. In contrast, for data with both contextual and parametric knowledge, we use \( \varepsilon = 1 \). Notably, only the linear layer $W$ is fine-tuning throughout the training phase.

\textbf{Inference.} At the inference stage, \( \varepsilon \) can adopt any value within the interval \([-1, 1]\). Specifically, an \( \varepsilon \) value of $-1$ corresponds to minimal reliance on parametric knowledge, whereas a value of $1$ indicates a strong inclination towards such knowledge. Detailed theoretical explanation on why HallE-Control works are elaborated upon in the Appendix.

\section{Experiment}
\begin{table*}
\centering
\captionsetup{aboveskip=2pt}\captionsetup{belowskip=0pt}
\caption{Comparison between baselines and the effect of indication on \textit{CCEval}. 'Only ind' means evaluation only on indicated objects; 'w/o ind' means evaluation without indicated objects; 'w/ ind' means evaluation with indicated objects.}
\label{tab:special_token}
\resizebox{13cm}{!}{
\begin{tabular}
{l|c@{\hspace{1.0\tabcolsep}}c@{\hspace{1.0\tabcolsep}}|c@{\hspace{1.0\tabcolsep}}c@{\hspace{1.0\tabcolsep}}c@{\hspace{1.0\tabcolsep}}|c@{\hspace{1.0\tabcolsep}}c@{\hspace{1.0\tabcolsep}}}
\toprule
\textbf{Setting} & \textbf{LLM} & \textbf{Resolution} &  \textbf{\textit{CHAIR}$_s$}$\downarrow$ & \textbf{\textit{CHAIR}$_i$}$\downarrow$ & \textbf{Coverage}$\uparrow$ & \textbf{Avg. Length}$\uparrow$ & \textbf{Avg. Object}$\uparrow$\\
\midrule
158K baseline & LLaVA$_\textsc{7B}$ & 224x & 82.00 & 25.30 & 33.58 & 109.89 & 9.31 \\
only ind. & LLaVA$_\textsc{7B}$ & 224x & 53.00 & 63.90 & 12.01 & -- & 1.66 \\
w/o ind. & LLaVA$_\textsc{7B}$ & 224x & 57.00 & 17.10 & 37.60 & 108.94 & 7.63\\ 
\rowcolor[gray]{.9}
w/ ind. & LLaVA$_\textsc{7B}$ & 224x &  57.00 & 14.00 & 37.62 & 108.94 & 9.22\\
\midrule
158K baseline & LLaVA Llama 2$_\textsc{13B}$ & 336x & 64.00 & 16.00 & 33.37 & 108.52 & 8.92 \\
only ind. & LLaVA Llama 2$_\textsc{13B}$ & 336x & 52.00 & 62.31 & 19.90 & -- & 1.3 \\
w/o ind. & LLaVA Llama 2$_\textsc{13B}$ & 336x & 52.00 & 11.62 & 34.70 & 106.94 & 7.23 \\
\rowcolor[gray]{.9}
w/ ind. & LLaVA Llama 2$_\textsc{13B}$ & 336x & 52.00 & 9.86 & 39.31 & 106.94 & 8.52 \\

\bottomrule
\end{tabular}
}
\end{table*}

\begin{table*}
\centering
\captionsetup{aboveskip=2pt}\captionsetup{belowskip=0pt}
\caption{Performance of \textit{HallE-Control}.}
\label{tab:switch}
\resizebox{12cm}{!}{
\begin{tabular}
{l|c@{\hspace{1.0\tabcolsep}}c@{\hspace{1.0\tabcolsep}}|c@{\hspace{1.0\tabcolsep}}c@{\hspace{1.0\tabcolsep}}c@{\hspace{1.0\tabcolsep}}|c@{\hspace{1.0\tabcolsep}}c@{\hspace{1.0\tabcolsep}}}
\toprule
\textbf{Control value} & \textbf{LLM} & \textbf{Resolution} &  \textbf{\textit{CHAIR}$_s$}$\downarrow$ & \textbf{\textit{CHAIR}$_i$}$\downarrow$ & \textbf{Coverage}$\uparrow$ & \textbf{Avg. Length}$\uparrow$ & \textbf{Avg. Object}$\uparrow$\\
\midrule
1 & LLaVA$_\textsc{7B}$ & 224x & 89.00 & 26.60 & 32.80 &108.54 & 9.72 \\
0.5 & LLaVA$_\textsc{7B}$ & 224x & 85.00 & 27.92 & 34.02 & 109.33 & 8.81 \\
-0.5 & LLaVA$_\textsc{7B}$ & 224x & 81.00 & 24.88 & 35.87 & 118.08 & 8.04 \\
\rowcolor[gray]{.9}
-1 & LLaVA$_\textsc{7B}$ & 224x & 76.00 & 20.90 & 33.88 & 133.79 & 8.02 \\
\midrule
1 & LLaVA Llama 2$_\textsc{13B}$ & 336x & 65.00 & 14.58 & 36.14 & 102.18 & 8.37 \\
0.5 & LLaVA Llama 2$_\textsc{13B}$ & 336x & 65.00 & 14.44 & 32.32 & 103.51 & 8.45 \\
-0.5 & LLaVA Llama 2$_\textsc{13B}$ & 336x & 66.00 & 13.79 & 33.07 & 105.57 & 8.41 \\
\rowcolor[gray]{.9}
-1 & LLaVA Llama 2$_\textsc{13B}$ & 336x & 43.00 & 6.37 & 34.37 & 136.28 & 8.79 \\
\bottomrule
\end{tabular}
}
\end{table*}

\subsection{Finetune on Parametric Joint Data}
Before we present experiments on HallE-Control, we show the upper-bound experiments on how well the model can indicate parametric knowledge. We directly finetune the LLaVA model on parametric joint data. Intuitively, the model is trained on data indicating parametric knowledge. Its output should identify parametric knowledge accurately. Specifically, the model should put a bracket around every "guessed" objects for indication of hallucination.

Therefore, we evaluate the object hallucination in three different settings:
1. Evaluation only on indicated objects: We do CCEval only on objects inside the bracket. The result should reflect a high level of hallucination.
2. Evaluation without indicated objects:
We disregard objects in bracket and calculate CCEval. The result should reflect a low level of hallucination.
3. Evaluation with indicated objects: We calculate CCEval on all objects. Due to modification of the settings, we slightly change definition of CHAIR scores in CCEval as detailed in Appendix.

\textbf{Evaluation only on indicated objects.} The hallucination level for indicated objects, $CHAIR_i$, is 63.90 for LLaVA$_{7B}$ and 62.31 for LLaVA$_{13B}$. It is considerably higher than the baselines all other models. Concurrently, their coverage is 12.01 and 19.09 for LLaVA$_{7B}$ and LLaVA$_{13B}$, respectively, which both of them significantly lower than the coverage of 33.58 for LLaVA$_{7B}$ baseline model. The experiment results show the object within special tokens has significantly higher hallucination rate and lower coverage rate which support our assumption that the object inside the indication tokens are objects from parametric knowledge.

\textbf{Evaluation without indicated objects.} For objects outside of the special token scope, we find hallucination is markedly reduced. CHAIR$_s$ decreased from 82 to 57 which is 30.5\% percent improvement; CHAIR$_i$ decrease 32.4\%, from 25.3 to 17.10, compared to the baseline. This suggests that the model is less prone to make erroneous assumptions for objects not marked by brackets. This is interesting because the model perfectly captures the intention of marking parametric objects in training data and replicate the behavior during inference.

\textbf{Evaluation with indicated objects.}: We observe a significant decline in the hallucination without any reduce in object coverage. LLaVA$_{7B}$ CHAIR$_i$ improved from 25.30 to 14.00 which has 44.66\% improvements. For LLaVA$_{13B}$ CHAIR$_i$ improved from 16 to 9.86 which also has 38.38\% improvements.

\subsection{Hallucination Controlling}

During model inference, we select 4 different $\varepsilon$
, ranging from $-1$ to $+1$. As shown in Table \ref{tab:switch}, we evaluate HallE-Control$_{7B}$ and HallE-Control$_{13B}$ model, which use LLaVA as backbone. For the 7B model, we train it by removing the special token in parametric joint data, showing that indication is not a necessary condition for control module to work. $\varepsilon=-1$ 
means the control module trained purely on contextual only data, which try to minimize the hallucination, where $\varepsilon=+1$ maximize parametric knowledge output. The results show that as $\varepsilon$ 
increase, CHAIR$_i$ 
increase from 20.90 to 26.6, the coverage keeps at a similar level. 

For the 13B model, we keep the indication inside the parametric joint data. The HallE-Control$_{13B}$ achieves the best in object existence hallucination metric. With control value set to -1 and indication, we have $\textsc{CHAIR}_s=43$ versus baseline's 64 and $\textsc{CHAIR}_i=6.37$ vs baseline's 16, and the coverage of the model is not decreasing.
\section{Related Work}
\paragraph{Large Multimodal Models (LMMs).}
The rapid advancements in Large Language Models (LLMs)~\citep{touvron2023llama,chung2022scaling,touvron2023llama2,anil2023palm,driess2023palm,scao2022bloom,OpenAI2023GPT4TR} combined with a surge in open-source initiatives, have paved the way for the emergence of extensive vision-language models~\citep{liu2023visual, balanced_vqa_v2, zhu2023minigpt,2023llavarlhf,ye2023mplug,bai2023qwen,chen2023shikra,peng2023kosmos}. LLaVA introduces the concept of integrating a simple projector during LLM fine-tuning. Chatspot~\citep{zhao2023chatspot} follow LLaVA's model structure, but embed region of interest into instruction data. GPT4RoI~\citep{zhang2023gpt4roi} and Shikra~\citep{chen2023shikra} add grounding tasks to LLaVA structure models. Instead of using detector to provide region information to the model, we use detector to filter objects for alignment between vision and language information. Additionally, BLIP2~\citep{li2023blip} and InstructBLIP~\citep{instructblip} present Q-former-based LMMs. Multimodal-GPT~\citep{gong2023multimodalgpt} and Otter~\citep{li2023otter} aims to improve OpenFlamingo's~\citep{alayracflamingo,awadalla2023openflamingo} directive adherence. mPLUG-Owl~\citep{ye2023mplug} suggests a two-step method: first train vision models, then refining the language model using techniques like LoRA. Our work utilize a linear layer to control object existence hallucination within LMMs.
\paragraph{Evaluation on LMMs.}
The evaluation of large vision-and-language models (LMMs)~\citep{yu2023mmvet,liu2023improvedllava,liu2023llava} is challenging due to the intricate nature of generation tasks they undertake. Some of the VQA-based benchmarks~\citep{VQA,hudson2018gqa,gurari2018vizwiz} require models to identify objects, colors, or quantities, while others~\citep{MMBench,li2023seedbench,lu2022learn} offer multiple-choice questions. POPE~\citep{Li-hallucination-2023} and MME~\citep{fu2023mme} evaluate object hallucination using paired yes/no questions on existence, color, counting, OCR, and etc. While VQA-based benchmarks are cheap and straightforward, we find them cannot accurately reflect object hallucination for detailed captions. 
Besides VQA benchmarks, ROUGE~\citep{lin-2004-rouge, elliott-keller-2014-comparing} use n-gram to evaluate similarity between ground truth and model inferences. CIDr~\citep{7299087} is a triplet-based method of collecting human annotations to measure consensus. CHAIR~\citep{objectHallucination} evaluates caption hallucination with object concepts. These methods are constraint by ground truth length or word variance and cannot clearly reflect hallucination with object coverage information. \cite{wang2023evaluation} uses a language model to predict whether the caption exist hallucination, which is cheaper than GPT-4. Our work introduces CCEval, including CHAIR metrics, object coverage, average sentence length and number of objects to overcome limitations of previous evaluations.

\paragraph{Hallucination}
Hallucinations~\citep{10.1145/3571730,shi2023replug, lin2021truthfulqa} have been widely studied in traditional NLG~\citep{ji2023survey} tasks, including machine translation~\citep{Zhou2020DetectingHC,lee2019hallucinations}, data-to-text~\citep{rebuffel2021controlling, kasner2022neural, lee2022factuality}, summarization~\citep{cao-etal-2022-hallucinated}, dialogue~\citep{dziri2022faithdial} and QA~\citep{shuster-etal-2021-retrieval-augmentation}. For LMMs, previous studies have been mainly focusing on object hallucination~\citep{marino2019ok, macleod2017understanding, li2023mimic, Li-hallucination-2023}. POPE~\citep{Li-hallucination-2023} reveals object existence hallucination may related with label distributions, such as object co-occurance. Earlier than POPE, \cite{9706727} balance object co-occurance to decrease hallucination. LRV~\citep{liu2023aligning} finds the cause of hallucination in VQA benchmarks, especially unbalanced answer distribution and lack of negation information. Our work raise another important cause: misalignment between the vision and language information captured by models. More interestingly, we can explain balancing labels in \cite{9706727} as trying to weaken the parametric knowledge caused by the misalignment.
\section{Conclusion}
In summary, this study delves deep into the object hallucination phenomena within the detailed captions of LMMs, advancing understanding of the accuracy and unwarranted inference in describing visual details. We introduce a novel and comprehensive evaluation method for object existence hallucination in detailed captions. We conduct an in-depth and component-wise analysis of LMMs, meticulously examining each element that might result in hallucination. We further identify an alignment issue between the vision encoder and the instruction data. To alleviate such hallucination, we introduce controlling parameters over LMMs to condition the inference of objects.

\bibliography{colm2024_conference}
\bibliographystyle{colm2024_conference}

\appendix
\section{Appendix}
\subsection{Why not all Hallucination Bad?}
In our study, we define "object existence hallucination" to be a phenomenon where a description makes reference to objects that is not present in the image. However, such hallucinations, when properly harnessed, can be regarded as instances of imagination. Human beings frequently use imagination to successfully accomplish tasks, often without even realizing it. Here, we present several scenarios in Table \ref{tab:case1benefit}, Table \ref{tab:case2benefit}, and Table \ref{tab:case3benefit} in which imagination and judicious inference prove to be beneficial for various downstream applications, including robotic manipulation and content moderation, among others. Our work successfully exercise control over the extent of generalization rather than attempting to eliminate all instances of imagined objects. More importantly, our model can indicate imagined objects with \texttt{[object]} marker.

\begin{table*}[!h]\centering
\begin{minipage}{1.0\columnwidth}\vspace{0mm}    \centering
\begin{tcolorbox} 
\centering
\footnotesize
\begin{tabular}{p{0.97\columnwidth} c}
\ArgSty{\bf First Case:} In the context of robotic applications, Large Multimodal Models (LMMs) encounter challenges when camera does not capture all objects needed. In such situations, it becomes crucial for a detailed caption to infer the surrounding objects. For instance, in this scenario, if the robot is ask for manipulating the chairs, it can successfully locate the chairs and notice people sitting on the chairs, even if they are not shown in the picture.

\\

\begin{tabular}{ p{6cm}  p{6cm}}
      \ArgSty{\bf Image:} & \ArgSty{\bf Caption:} \\ 
     \raisebox{-\totalheight}{\includegraphics[width=0.33\textwidth]{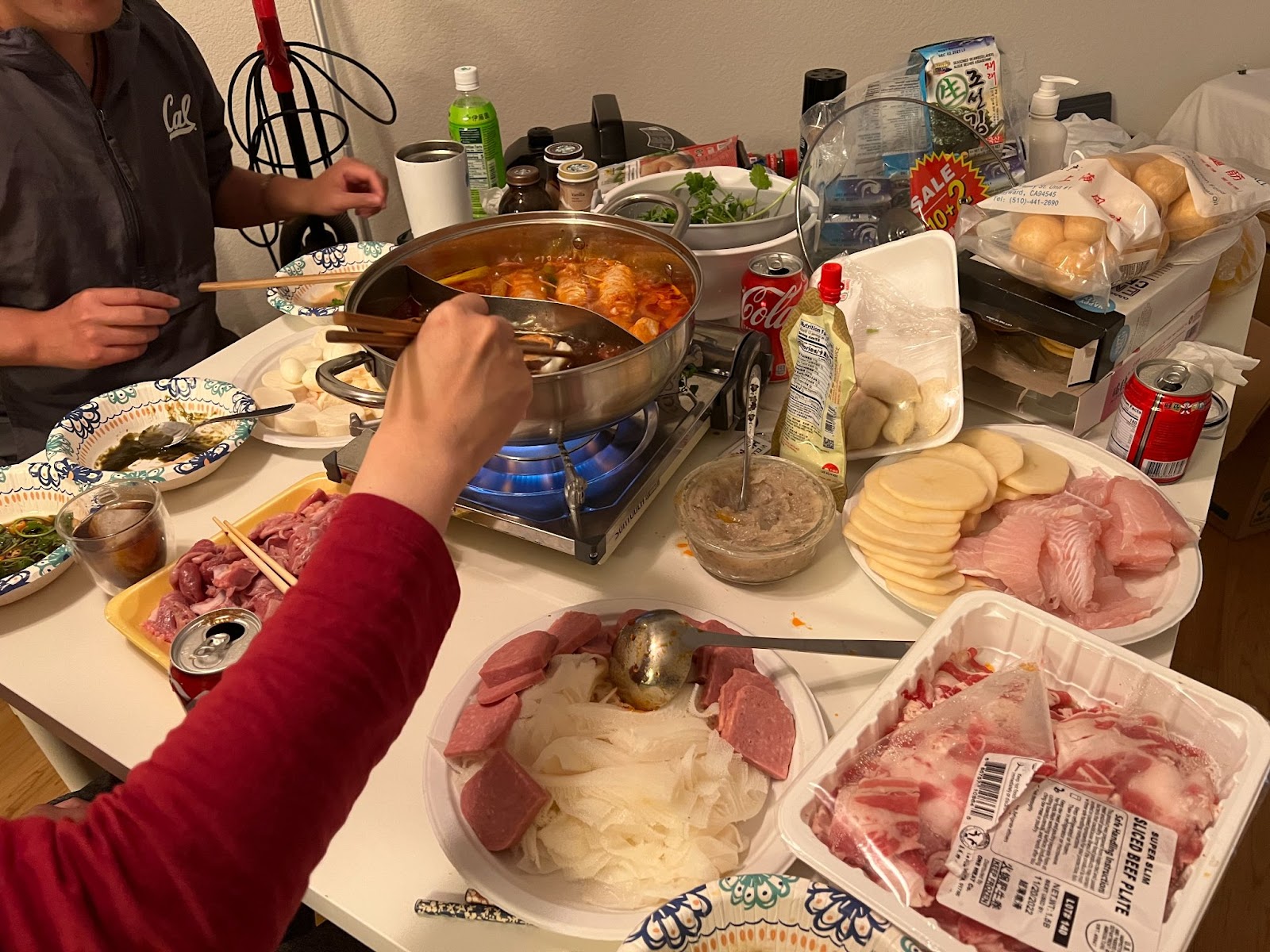}}
      & 
      The image depicts a group of people gathered around a dining table, preparing a meal together. The table is filled with various food items, including several bowls containing different dishes, and a few plates with different types of food. There are also multiple bottles and cups scattered across the table, likely containing beverages or condiments.
Two \colorbox{hotpink}{people} can be seen in the scene, one on the left side and another on the right side of the table. They are actively engaged in the cooking process, with one person stirring a pot of food on the stove. The table is surrounded by \colorbox{hotpink}{chairs}, with one on the left side and another on the right side.
In addition to the food and people, there are a few utensils on the table, such as a spoon and a \colorbox{hotpink}{knife}. A \colorbox{hotpink}{cell phone} can also be seen on the table, possibly being used to look up recipes or communicate with others.

    \end{tabular}

\end{tabular}
\end{tcolorbox}
\vspace{-2mm}
\caption{Beneficial hallucination in robotic applications. Caption is from LLaVA1.5. Imagined objects are \colorbox{hotpink}{highlighted}.}
\label{tab:case1benefit}
\end{minipage}
\end{table*}

\newpage
\BgThispage
\begin{table*}[!h]\centering
\begin{minipage}{1.0\columnwidth}\vspace{0mm}    \centering
\begin{tcolorbox} 
\centering
\footnotesize
\begin{tabular}{p{0.97\columnwidth} c}
\ArgSty{\bf Second Case:} In the context of content moderation, LMMs need to infer objects to recognize cases of violation. For instance, in this scenario, a man is not explicitly present in the image but can be correctly inferred to help moderation.

\\

\begin{tabular}{ p{6cm}  p{6cm}}
      \ArgSty{\bf Image:} & \ArgSty{\bf Caption:} \\ 
     \raisebox{-\totalheight}{\includegraphics[width=0.3\textwidth]{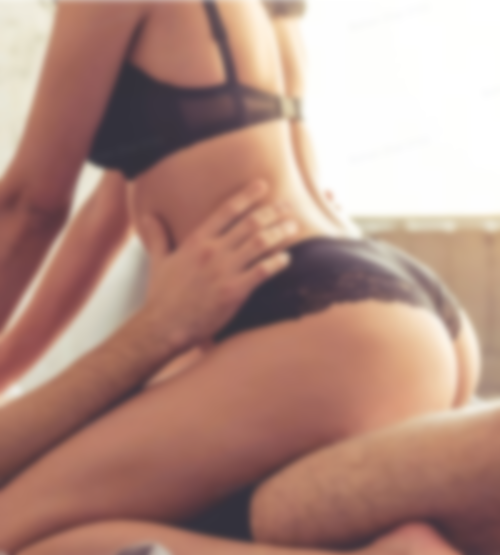}}
      & 
      The image features a \colorbox{hotpink}{man} and a woman lying on a bed, with the woman sitting on top of the man. The woman is wearing black lingerie, and the man has his hands on her hips, possibly holding her in place. The couple appears to be enjoying a moment of intimacy together.
The bed occupies most of the image, with the couple being the main focus. The woman's black lingerie is prominently visible, while the man's hands are placed on her hips, emphasizing the close and intimate nature of their interaction.

    \end{tabular}

\end{tabular}
\end{tcolorbox}
\vspace{-2mm}
\caption{Beneficial hallucination in content moderation. Caption is from LLaVA1.5. Imagined objects are \colorbox{hotpink}{highlighted}.}
\label{tab:case2benefit}
\end{minipage}
\end{table*}

\begin{table*}[!h]\centering
\begin{minipage}{1.0\columnwidth}\vspace{0mm}    \centering
\begin{tcolorbox} 
\centering
\footnotesize
\begin{tabular}{p{0.97\columnwidth} c}
\ArgSty{\bf Third Case:} In the context of robotic applications, LMMs encounter challenges when dealing with poorly captured images. These images often feature partially occluded or blurred objects. In such situations, it becomes crucial for a detailed caption to go beyond what is visible and accurately infer the presence of occluded objects. This capability is essential to address safety considerations. For instance, in this case, there is a cat behind the door reaching out its paw to play with the black ribbon, but the body of the cat is not explicitly shown. If the model can successfully recognize a cat, the robot should refrain from pushing the door, thus ensuring the safety of both the robot and the cat.

\\

\begin{tabular}{ p{6cm}  p{6cm}}
      \ArgSty{\bf Image:} & \ArgSty{\bf Caption:} \\ 
     \raisebox{-\totalheight}{\includegraphics[width=0.33\textwidth]{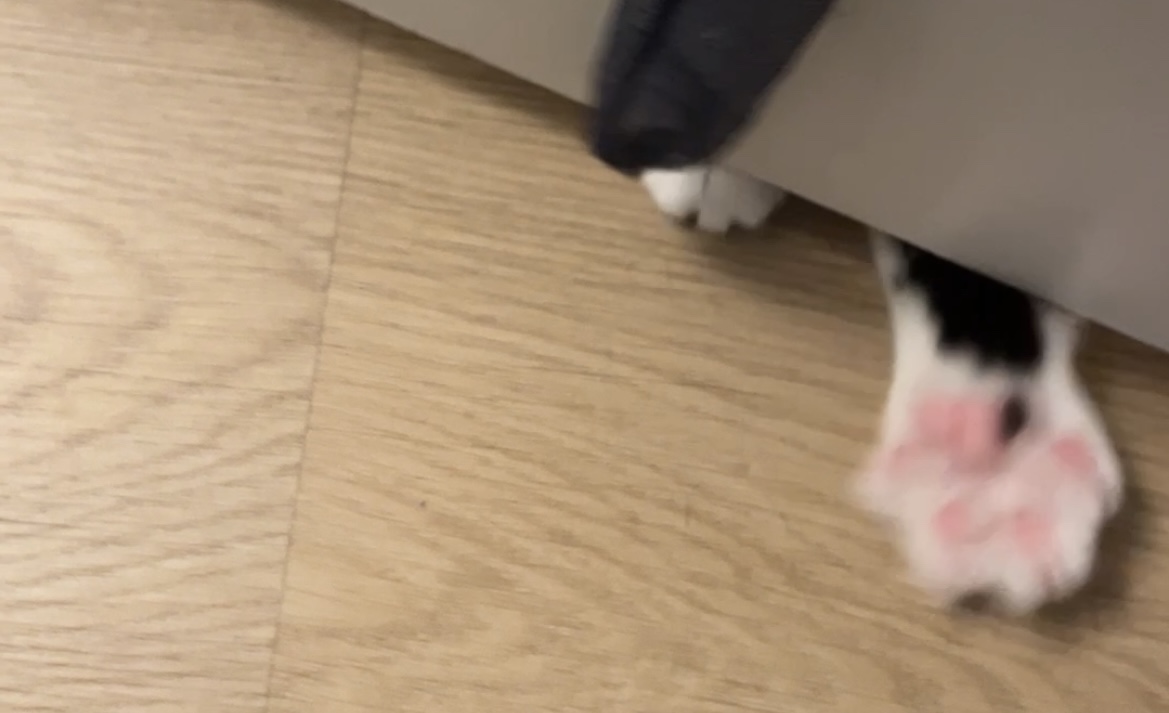}}
      & 
      The image features a black and white \colorbox{hotpink}{cat} with pink paws, standing on a wooden floor and looking under a door. The cat appears to be curious about what is happening behind the door. The door is positioned on the left side of the image, taking up a significant portion of the background. The cat is located towards the right side of the image, with its paws visible as it peers under the door.
    \end{tabular}

\end{tabular}
\end{tcolorbox}
\vspace{-2mm}
\caption{Beneficial hallucination in robotic applications. Caption is from LLaVA1.5. Imagined objects are \colorbox{hotpink}{highlighted}.}
\label{tab:case3benefit}
\end{minipage}
\end{table*}

\subsection{More Details on Analyzed Models}

In Section 2, we conduct analysis on different LMMs. Here, we provide more details of each model:

\textbf{LLaVA} uses a linear projector to map visual token as a soft-prompt into LLM input tokens. LLaVA has a two-stage training, where the initial stage focuses on simple caption pretraining solely for the linear projector, while the subsequent stage finetunes both the projector and LLM on instruction data. Instruction data leverages language-only GPT-4 by inputting visual ground truth from COCO dataset.

\textbf{InstructBLIP} adopts the BLIP-2 architecture, and is distinguished by its training of a Q-former, which bridges the frozen vision encoder and LLM. InstructBLIP's instruction fine-tuning spans across 26 distinct datasets.

\textbf{Shikra} mirrors LLaVA's model structure. It eliminates the pretrain stage, but introduce grounding task during finetuning. Shikra is trained on multiple datasets like InstructBLIP.

\subsection{Training Data Quality}



We sampled three images from the MSCOCO dataset, as illustrated in Table \ref{tab:data_quality}. For each image, we present the visual content, a detailed caption generated by GPT-4 based on bounding boxes and regional captions, and the object ground truth labels derived from the MSCOCO dataset annotations.

\textbf{First Image.} This image showcases three remote controls, posing a unique challenge. The ambiguity lies in distinguishing the type of remote, be it for video games or televisions. Additionally, the remotes near the wall are relatively small, making them harder to see. A person is partially visible in the image, with only their knees being evident. There is also an incomplete bottle on the image's left side.

\textbf{Second Image.} A significant issue with this image is the individuals visible behind a window. Not only detection models struggle to recognize them, even for human observers, counting the individuals inside the train is challenging. This particular issue is prevalent in many of MSCOCO's traffic-related images.

\textbf{Third Image.} This image depicts a table around which two individuals are seated. Close observation is required to recognize both individuals. A clear indication of one person is a pair of hands, while the other individual is considerably harder to spot. Additionally, there seems to be an annotation error in the ground truth labels: it indicates only one bowl, neglecting to include plates and other items. Upon closer inspection, there are four plates.

\begin{table}[h!]
     \begin{center}
     \begin{tabular}{ p{6cm}  p{6cm}  p{2cm}  }
     \toprule
      Image & Caption & GT Labels \\ 
    \cmidrule(r){1-1}\cmidrule(lr){2-2}\cmidrule(l){3-3}
     \raisebox{-\totalheight}{\includegraphics[width=0.33\textwidth, height=40mm]{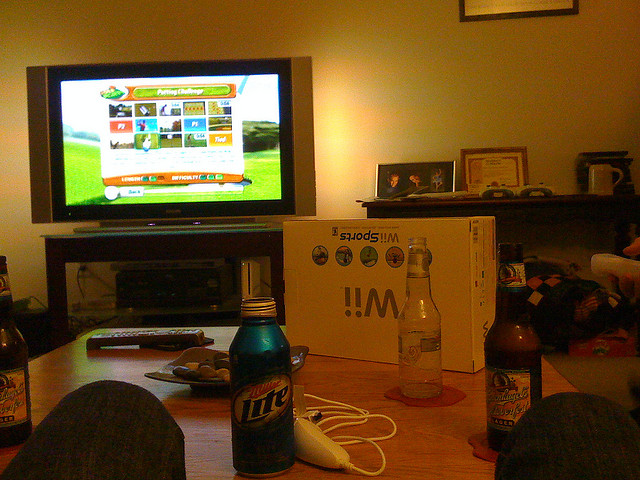}}
      & 
      The image depicts a cozy entertainment room featuring a flat screen TV situated against the wall. A dining table can be seen next to the TV, with some bottles and a cup placed on top of it. \textcolor{hotpink}{Multiple video game controllers}, including \textcolor{hotpink}{Wii remotes}, are also present on the dining table, suggesting that people are using the TV to play video games. To the far right of the scene, \textcolor{hotpink}{another remote} is located close to the wall.
      
    \textcolor{hotpink}{A person} appears to be sitting or standing behind the dining table, likely either watching the game or waiting for their turn to play. The assortment of bottles on the table suggests that the guests are enjoying drinks during their video game session.
      & 
      tv: 1
      
      cup: 1
      
      remote: 3
      
      bottle: 4
      
      person: 1
      
      dining table: 1
      
      \\ \midrule

     \raisebox{-\totalheight}{\includegraphics[width=0.33\textwidth, height=40mm]{figures/000000157923.jpg}}
      & 
In the image, an orange mass transit trolley is making its way through a city. A person is crossing the street in front of the trolley while holding a garbage bag, appearing to be cautious about the approaching vehicle. \textcolor{hotpink}{Another person} is standing close to the person crossing the street, and there are \textcolor{hotpink}{two more individuals} nearby. 

In the scene, various vehicles surround the trolley, including cars, buses, and trucks. One of the cars is parked right behind the trolley, while another is situated farther back. Two buses can be seen, with one staying behind the trolley and the other on the right side of it. A truck is also present at the far left side of the scene.

A woman nearby is holding a handbag, completing the busy urban setting.
      & 
      car: 3
      
      bus: 2
      
      truck: 1
      
      train: 1
      
      person: 5
      
      handbag: 1
      
      \\ \midrule

     \raisebox{-\totalheight}{\includegraphics[width=0.33\textwidth, height=40mm]{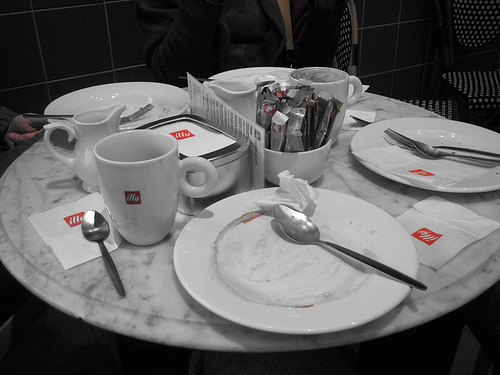}}
      & 
The image shows a table in a restaurant, which appears to have been recently used for a meal. The table is set with four place settings, including white dishes. On the table, there are multiple cups, a bowl, and a mix of silverware like forks and spoons laid out. Some of the dishes look dirty, with used napkins and eating utensils scattered around.

In the background, \textcolor{hotpink}{two people} can be seen, one situated on the left side and another on the right side behind the table. There are also \textcolor{hotpink}{chairs} located near the table, with one \textcolor{hotpink}{chair} positioned close to the left side and another \textcolor{hotpink}{chair} closer to the right side of the frame.
      & 
      cup: 5
      
      spoon: 3
      
      fork: 2
      
      person: 2
      
      chair: 2
      
      bowl: 1

      dining table: 1
      
      \\ \bottomrule

      \end{tabular}
      \caption{Quality of training data.}
      \label{tab:data_quality}
      \end{center}
      \end{table}

\subsection{CCEval}
In this paper, we introduce CCEval, a GPT-4 assisted $\textbf{C}$aption  $\textbf{C}$oncept Matching $\textbf{Eval}$uation. CCEval provides concept matching method and metric that can be extended to larger scale. Here, we detail the process of evaluation:

\textbf{Step 0.} Collect ground truth objects from annotations in a list.

\textbf{Step 1.} Run the model through the evaluation set of 100 images from Visual Genome. Specifically, prompt the model with "Describe the image in detail" and obtain 100 captions as model outputs.

\textbf{Step 2.} Extract objects mentioned in 100 captions using GPT-4.

\textbf{Step 3.} For each caption, match the extracted objects with ground truth objects using GPT-4.

GPT-4 prompts are provided in the next section. After Step 3, we get the following information: matched objects (no hallucination), unmatched objects (hallucination), total objects mentioned, total ground truth objects. Our metric calculates:

$$\textsc{CHAIR}_i = \frac{|\{\text{unmatched\ objects}\}|}{|\{\text{total\ objects\ mentioned}\}|}$$
$$\textsc{CHAIR}_s = \frac{|\{ \text{sentences\ with\ unmatched\ objects\}}|}{|\{\text{total\ sentences\}}|}$$
$$\text{Coverage} = \frac{|\{ \text{matched\ objects\}}|}{|\{\text{total\ ground\ truth\ objects\}}|}$$

$$\text{Average Length} = \frac{|\{ \text{total\ words\}}|}{|\{\text{number of examples}\}|}$$
$$\text{Average Objects} = \frac{|\{ \text{total\ objects\ mentioned\}}|}{|\{\text{number of examples}\}|}$$

Note that we do not strictly enforce consistent constraints (same coverage, average length, and average objects) across models, neither this is realistic to achieve. Although our intention is to make the model comparable and push the model to express more objects, it is sometimes not possible for some of the models, such as BLIP2, to express detailed captions, and our coverage, average length, and average number of object metrics in CCEval can accurately reflect if any model has such limitations.

In the main text, "consistent constraints'' are mainly achieved through prompting. We use the same prompt - “describe this image in detail” - across all four models in Table \ref{tab:our_eval}. Simply prompting can result in similar length and number of objects captions in this case since the instruction finetuning data for detailed captioning all comes from a similar source (LLaVA synthetic data). For models with different instruction finetuning data, we suggest further adjustment of the prompt or set the min\_length and max\_length in the inference function.

\subsection{Prompt}
 \label{sec:prompt}

Table \ref{tab:prompt1} is the prompt we use for caption object extraction. 

Table \ref{tab:prompt2} is the prompt we use for hallucination object matching.

Table \ref{tab:prompt3} is the prompt we use for finding ground truth object coverage.
\begin{table*}[!h]\centering
\begin{minipage}{1.0\columnwidth}\vspace{0mm}    \centering
\begin{tcolorbox} 
\centering
\footnotesize
\begin{tabular}{p{0.97\columnwidth} c}
\ArgSty{\bf Caption Object Extract Prompt:}

\\
\textbf{User:} \\
I have a description of an image, and I want to get objects from this description and return these objects in a list the object should be a noun, and I don't want duplicated objects.
I don't want the scene name to be included, such as some caption describing the image as a scene or depicting a position or a situation or place, this thing is not an object, and doesn't need to be included.
Here some objects are inside [] which we want to ignore.
Here are some examples: \\
\ArgSty{\bf Example 1:} \\
\textbf{Input:} \\
caption = "The image features a bathroom sink situated under a large mirror. The sink is accompanied by a soap dispenser, and there are multiple toothbrushes placed around it. A few cups can be seen scattered around the sink area as well. $\backslash$n $\backslash$n In addition to the sink, there is a toilet visible to the left side of the bathroom. The overall scene gives an impression of a well-equipped and functional bathroom space. Also a [brush] can been seen." \\

\textbf{Answer:} \\
objects = ['sink', 'mirror', 'soap dispenser', 'toothbrush', 'cup', 'toilet']
\\
Here we can see [brush] is ignored because its inside []. bathroom is the place not object, so not included.
\\
\\
\ArgSty{\bf Example 2:} \\ 
\textbf{Input:} \\
caption = "The image depicts a cluttered dining room with a large kitchen table in the center. The table is covered with dirty dishes, including plates, bowls, cups, and utensils. There are several chairs around the table, with some placed closer to the center and others positioned at the edges.  In addition to the dishes, there is an apple sitting on the table, likely left over from a meal or snack. A bottle of water can be seen on the table as well, and a [flower], adding to the messy atmosphere of the room." \\
\\
\textbf{Answer:} \\
objects = ['table', 'dish', 'bowl', 'cup', 'utensil', 'chair', 'apple', 'water'] \\
\\
Here [flower] is in [], should be ignored. Here dining room and room are places, so ignored, not in objects. \\
\\
\ArgSty{\bf Example 3:} \\ 
\textbf{Input:} \\
caption = "The image depicts a busy city street with a pedestrian crossing in a sunny day. A man is walking across the street, carrying a backpack and wearing a jacket." \\
\\
\textbf{Answer:} \\
objects = ['street', 'pedestrian crossing', 'man', 'backpack', 'jacket'] \\ 
\\
Here 'city' is a place, so not an object so not included in objects. 'The image depicts' is about the image caption task, so not an object in the scene. 'sunny' or 'sunny day' or 'day' are not objects in the image, this is a time situation so not an object, can't in objects. \\
\\
\ArgSty{\bf Example 4:} \\ 
\textbf{Input:} \\
caption = "The image depicts an office cubicle with a desk in the center. The desk is equipped with a computer, a keyboard, and a mouse." \\
\textbf{Answer:} \\
objects = ['desk', 'computer', 'keyboard', 'mouse'] \\
\\
Here the office is a place so not in objects. Here 'center' is not object, 'center' is position, not an object, same thing like 'left' or 'right' etc. \\
\\
\ArgSty{\bf Inputs:} \\ 
caption = {cap} \\
\textbf{Answer:} \\
objects = 
\\
\end{tabular}
\end{tcolorbox}
\vspace{-2mm}
\caption{Caption object extract prompt}
\label{tab:prompt1}
\end{minipage}
\end{table*}

\begin{table*}[!h]\centering
\begin{minipage}{1.0\columnwidth}\vspace{0mm}    \centering
\begin{tcolorbox} 
\centering
\footnotesize
\begin{tabular}{p{0.97\columnwidth} c}
\ArgSty{\bf Hallucination Prompt:}

\\
\textbf{User:} \\
I have two lists of objects, list\_A, and list\_B, I want to return a list hallucination which finds items in list\_B don't appear in list\_A, sometimes same object can be expressed in different ways in list\_A and list\_B, we treat different expression but similar meaning objects as matched, not include in mismatch list. \\

\\
\ArgSty{\bf Example 1:} \\
\textbf{Input:} \\
list\_A = ['reflection of light', 'view of office building', 'street chair', 'white car', 'red car', 'dark hair', 'bagpack', 'black shoes', 'dark pants', 'bikes', 'street', 'street light'] \\
list\_B = ['two cars', 'dark bagpack', 'yellow jacket', 'light', 'brick building', 'wood chair', 'chair', 'green car', 'dining room table', 'bike', 'city street', 'traffic light', 'sedan'] \\
\textbf{Answer:} \\
In this example, 'two cars' is just object 'car', we don't care about the number of object. Although 'bikes' and 'bike' is not the same word, but we treat singular nouns and plural nouns as the same thing, so it's not mismatch.
Here in list\_A's 'street' and list\_B's 'city street' are not exactly match but actually, city street can been seen as a kind of street, since city street is still a street, just in city,  so they are similar meaning, we don't treat it as a mismatch, even 'city street' seems more specific, but we only still treat it as a match not hallucination. Although there is 'street light' in list\_A but 'traffic light' is a different object, 'light' and 'street light' are aiming for providing lights, but 'traffic light''s purpose is providing signal, so they are different object. 'Sedan' is a different kind of car, so 'sedan' match 'car'. \\
\\
hallucination = ['yellow jacket', 'dining room table', 'traffic light'] \\

\\
\ArgSty{\bf Example 2:} \\ 
\textbf{Input:} \\
list\_A = ['bag', 'cloth', 'boy', 'Drinking glasses', 'table'] \\
list\_B =['backpack', 'jacket', 'young man', 'cup', 'kitchen table'] \\
\\
\textbf{Answer:} \\
In this example, 'bag' in list\_A and 'backpack' in list\_B have similar meaning, 'jacket' in list\_B can be seen as a kind of 'cloth' in list\_A still matching, and 'Drinking glasses' is kind of cup. In list\_B 'kitchen table' is a kind of table as 
'table' in list\_A so there is no hallucination. \\
\\
hallucination = [] \\

& \\
\ArgSty{\bf Example 3:} \\ 
\textbf{Input:}
list\_A = ['keyboard', 'mouse', 'moniter', 'cpu'] \\
list\_B = ['computer'] \\
\\
\textbf{Answer:} \\
Based on the objects, 'keyboard', 'mouse', 'moniter', 'cpu' they are all parts of a computer and they all appreared in list\_A, and list\_B's 'computer' is just a summary of all these objects, so there is no hallucination. \\
\\
hallucination = [] \\
\\
\ArgSty{\bf Inputs:} \\ 
list\_A = \{gt\} \\
list\_B = \{cap\_obj\} \\
\textbf{Answer:} \\
hallucination = 
\\
\end{tabular}
\end{tcolorbox}
\vspace{-2mm}
\caption{Object hallucination matching prompt}
\label{tab:prompt2}
\end{minipage}
\end{table*}

\begin{table*}[!h]\centering
\begin{minipage}{1.0\columnwidth}\vspace{0mm}    \centering
\begin{tcolorbox} 
\centering
\footnotesize
\begin{tabular}{p{0.97\columnwidth} c}
\ArgSty{\bf Coverage Prompt:}

\\
\textbf{User:} \\
I have two list of objects, list\_A and list\_B, I want to return a list named uncover which find items in list\_B doesn't appear in list\_A, sometimes same object can be expressed in different ways in list\_A and list\_B, we treat different expression but similar meaning objects as matched, not include in mismatch list.\\

\ArgSty{\bf Example 1:} \\
\textbf{Input:} \\
list\_A = ['two cars', 'dark bagpack', 'yellow jacket', 'light', 'brick building', 'wood chair', 'chair', 'green car', 'dining room table', 'bike', 'city street', 'traffic light', 'sedan']
list\_B = ['reflection of light', 'view of office building', 'street chair', 'white car', 'red car', 'dark hair']
\textbf{Answer:} \\
uncover = ['reflection of light', 'dark hair'] \\
\\
In this example \\
'reflection of light' cannot find matched object in list\_A, especially, 'light' is not equal to 'reflection of light'. \\
'view of office building' in list\_B can find matched object 'brick building' although they are not exactly same but they point to similar object. \\
'street chair' in list\_B can find 'chair', 'wood chair' in list\_A which is an alternate expression of 'chair'. \\
'white car' in list\_B can find 'two cars' in list\_A. \\
'red car' in list\_B can find 'two cars' in list\_A. \\
'dark hair' in list\_B cannot find anything similar in list\_A \\
\\
\ArgSty{\bf Example 2:} \\ 
\textbf{Input:} \\
list\_A = ['bag', 'cloth', 'boy', 'Drinking glasses', 'table'] \\
list\_B =['backpack', 'jacket', 'young man', 'cup', 'kitchen table', 'plate', 'apple'] \\
\\
\textbf{Answer:} \\
uncover = ['plate', 'apple'] \\
In this example, \\
'backpack' in list\_B can find 'bag' in list\_A has similar meaning, matched. \\
'jacket' in list\_B can be seen as a kind of 'cloth' in list\_A still matching; \\
'young man' in list\_B can match 'boy' in list\_A; \\
'cup' in list\_B is similar to 'Drinking glasses' in list\_A; \\
'kitchen table' is a kind of table as 'table' in list\_A so there is no uncovered items. \\
'plate' in list\_B but no object has same or similar meaning in list\_A. \\
'apple' in list\_B but no object has same or similar meaning in list\_A. \\
\\
\ArgSty{\bf Example 3:} \\ 
\textbf{Input:}
list\_A = ['keyboard', 'mouse', 'moniter', 'cpu'] \\
list\_B = ['computer'] \\
\\
\textbf{Answer:} \\
uncover = []
'computer' in list\_B can find 'keyboard', 'mouse', 'moniter', 'cpu' as whole thing in list\_A, matched. \\ 
\\
\ArgSty{\bf Inputs:} \\ 
list\_A = \{cap\_obj\} \\
list\_B = \{gt\} \\
\textbf{Answer:} \\
hallucination = 
\\
\end{tabular}
\end{tcolorbox}
\vspace{-2mm}
\caption{Object coverage prompt}
\label{tab:prompt3}
\end{minipage}
\end{table*}

\subsection{Sliding Window}
We pad the original image to dimensions of \(672 \times 672\). Then, we divide the image into a \(3 \times 3\) grid, where each cell measures \(224 \times 224\). The encoder processes these cells sequentially, starting from the top-left and moving towards the bottom-right. The visual tokens from each cell are then concatenated.

\subsection{Experiment Settings}
The LLaVA we used in all experiments is pretrained on on LCS-558k which is subset of LAION~\citep{schuhmann2021laion}, CC~\citep{sharma-etal-2018-conceptual} and SBU~\citep{Ordonez:2011:im2text} data, and finetuned on Instruct-158K instruction data. We use Vicuna version 1.3 as initialized language decoder and CLIP-Large as vision encoder. The 158K finetuning data consists of detailed caption, complex reasoning, and conversation.

We employed the RAM detector, specifically the RAM-14M variant, which uses a Swin-Large backbone. In the data generation stage, we focused on the 'detailed caption 23K' file from the LLaVA Instruction set comprising 158K entries. This file was generated using a specific prompt provided by LLaVA repository\footnote{\url{https://github.com/haotian-liu/LLaVA/tree/main/playground/data/prompts/detail_description}}. Our preprocessing involved adding brackets '[]' around objects filtered by the RAM detector to uniquely identify them in the captions.

Hallucination control only involves only finetuning a linear layer added before $lm\_head$ layer. We freeze all other layers and only finetune control layer with 3 epochs. The dataset is generated data with associated $\varepsilon$ value. We have 10K contextual only detailed caption data and 23K parametric joint detailed caption data. 

\subsection{More Experiments on LLaVA 1.5}
\begin{table*}
\centering
\captionsetup{aboveskip=2pt}\captionsetup{belowskip=0pt}
\caption{Experiments on LLaVA 1.5 on CCEval.}
\label{tab:llava1.5}
\resizebox{13cm}{!}{
\begin{tabular}
{l|c@{\hspace{1.0\tabcolsep}}c@{\hspace{1.0\tabcolsep}}|c@{\hspace{1.0\tabcolsep}}c@{\hspace{1.0\tabcolsep}}c@{\hspace{1.0\tabcolsep}}|c@{\hspace{1.0\tabcolsep}}c@{\hspace{1.0\tabcolsep}}}
\toprule
\textbf{Setting} & \textbf{LLM} & \textbf{Resolution} &  \textbf{\textit{CHAIR}$_s$}$\downarrow$ & \textbf{\textit{CHAIR}$_i$}$\downarrow$ & \textbf{Coverage}$\uparrow$ & \textbf{Avg. Length}$\uparrow$ & \textbf{Avg. Object}$\uparrow$\\
\midrule
665K baseline & LLaVA$_\textsc{13B}$ & 336x & 66.00 & 17.84 & 34.43 & 103.63 & 8.35 \\
665K baseline & LLaVA$_\textsc{13B}$ & 336x (SW) & 64.00 & 15.54 & 34.39 & 101.33 & 8.30\\ 
w/o ind. & LLaVA$_\textsc{13B}$ & 336x & 54.00 & 11.24 & 34.00 & 103.62 & 6.76\\ 
\rowcolor[gray]{.9}
w/ ind. & LLaVA$_\textsc{13B}$ & 336x &  54.00 & 8.91 & 35.94 & 103.62 & 8.53\\

\bottomrule
\end{tabular}
}
\end{table*}

The experiments on the main text are based on LLaVA 1.3, we provide experiments on LLaVA 1.5 in Table \ref{tab:llava1.5}. The results are consistent with the conclusion in the main text and hallucination is less in this new version.

\subsection{Modified CCEval of Section 4.1}
1. Evaluation only on indicated objects
$$\textsc{CHAIR}_i = \frac{|\{\text{hallucinated\ objects\_w/\_indication}\}|}{|\{\text{all objects\ w/\_indication mentioned}\}|}$$

$$\textsc{CHAIR}_s = \frac{|\{ \text{sentences\ with\ hallucinated\ object\_w/\_indication\}}|}{|\{\text{all\ sentences\ w/\_indication}|}$$
2. Evaluation without indicated objects
$$\textsc{CHAIR}_i = \frac{|\{\text{hallucinated\ objects\_w/o\_indication}\}|}{|\{\text{all objects\ w/o\_indication mentioned}\}|}$$

$$\textsc{CHAIR}_s = \frac{|\{ \text{sentences\ with\ hallucinated\ object\_w/o\_indication\}}|}{|\{\text{all\ sentences\ w/o\_indication}|}$$
3. Evaluation with indicated objects
$$\textsc{CHAIR}_i = \frac{|\{\text{hallucinated\ objects\_w/o\_indication}\}|}{|\{\text{all\ objects\ mentioned}\}|}$$

$$\textsc{CHAIR}_s = \frac{|\{ \text{sentences\ with\ hallucinated\ object\_w/o\_indication\}}|}{|\{\text{all\ sentences\}}|}$$

\subsection{Qualitative Results of HallE-Control}
This section shows qualitative results for HallE-Control in Table \ref{tab:vis1}, Table \ref{tab:vis2}, Table \ref{tab:vis3}, Table \ref{tab:vis4}, and Table \ref{tab:vis5}. $\varepsilon = -1$ means no imagination and $\varepsilon = 1$ means max imagination. Both models can output indication.

\begin{table*}[!h]\centering
\begin{minipage}{1.0\columnwidth}\vspace{0mm}    \centering
\begin{tcolorbox} 
\centering
\footnotesize
\begin{tabular}{p{0.97\columnwidth} c}
\ArgSty{\bf First Case} 

\\

\begin{tabular}{ p{6cm}  p{6cm}}
      \ArgSty{\bf Image:} & \ArgSty{\bf Caption with $\varepsilon = -1$:} \\ 
     \raisebox{-\totalheight}{\includegraphics[width=0.33\textwidth]{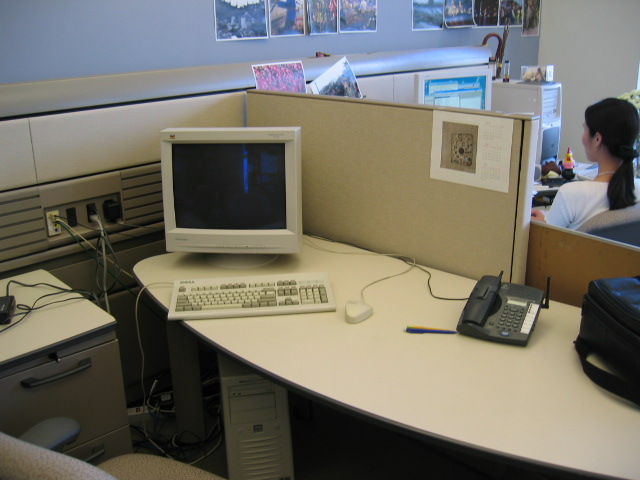}}
      & 
      The image features an office cubicle with a white desk and a computer setup. On the desk, there is a desktop computer with a monitor, keyboard, and mouse. A telephone is also present, indicating that this is a workstation with multiple functions. In addition to the main equipment, there are several books scattered around the workspace, suggesting that the user may be engaged in reading or research. A chair is positioned near the desk, ready for use, and a TV can be seen in the background, possibly for entertainment or information purposes during breaks. \\
      \\
      & \ArgSty{\bf Caption with $\varepsilon = 1$:} \\ 
      & The scene is set in an office environment, with a computer [mouse] and keyboard prominently placed on a desk. The keyboard is positioned towards the left side of the desk, while the [mouse] is located more towards the right. A computer monitor is situated on the desk, slightly behind the keyboard, indicating a typical workstation setup. A telephone is also present on the desk, located towards the right side, suggesting that the workspace is used for communication and computer-related tasks. In addition to these items, there are two [cups] on the desk, one slightly to the left of the keyboard and the other a bit further to the right. These [cups] might be used for holding beverages or storing stationery. The office is furnished with two chairs, one on the left side of the desk and the other on the right. Both chairs are positioned to face the desk, indicating that they are used for seating while working or engaging in phone conversations. The scene also features two [tv] screens, one on the left side of the room and the other on the right. These [tv] screens might be used for displaying relevant information or providing entertainment during work breaks.
        
    \end{tabular}

\end{tabular}
\end{tcolorbox}
\vspace{-2mm}
\caption{Example captions generated with HallE-Control$_{13B}$.}
\label{tab:vis1}
\end{minipage}
\end{table*}

\begin{table*}[!h]\centering
\begin{minipage}{1.0\columnwidth}\vspace{0mm}    \centering
\begin{tcolorbox} 
\centering
\footnotesize
\begin{tabular}{p{0.97\columnwidth} c}
\ArgSty{\bf Second Case} 

\\

\begin{tabular}{ p{6cm}  p{6cm}}
      \ArgSty{\bf Image:} & \ArgSty{\bf Caption with $\varepsilon = -1$:} \\ 
     \raisebox{-\totalheight}{\includegraphics[width=0.33\textwidth]{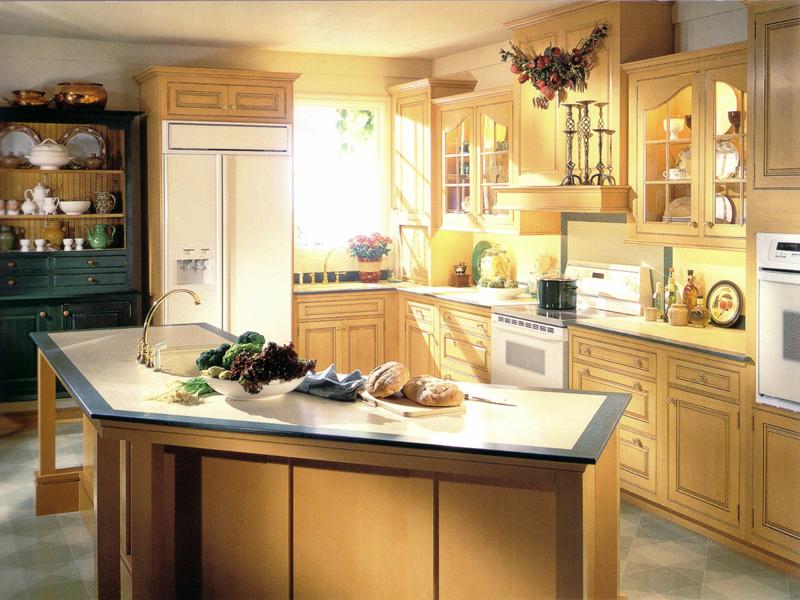}}
      & 
      The image depicts a large, clean, and well-organized kitchen with wooden cabinets and white countertops. The kitchen features a center island with various items placed on top of it, such as a knife, a loaf of bread, and some vegetables. There are multiple [bottles] and cups scattered around the kitchen, as well as a vase on the counter. In addition to the island, the kitchen is equipped with a refrigerator, a microwave, and two ovens, ensuring that it is well-equipped for cooking and food preparation. The presence of several vases and potted plants adds a touch of greenery and decoration to the space, making it inviting and pleasant.\\
      \\
      & \ArgSty{\bf Caption with $\varepsilon = 1$:} \\ 
      & The scene is a large, well-lit kitchen with a center island. The island is covered in various food items, including [oranges], [apples], [bottles], [cups], [spoons], [knives], [bowls], [bananas], [broccoli], [potted plants], [vases], [bread], [cake], [wine glasses], [potted herbs], [potted flowers], [books], a [clock], a [potted plant] on the counter, a [vase] on the counter, a [bottle] on the counter, a [spoon] on the counter, a [knife] on the counter, a [bowl] on the counter, a [banana] on the counter, a [broccoli] on the counter, a [wine glass] on the counter, a [cup] on the counter, a [potted herb] in the background, a [potted flower] in the background, a [book] in the background, a [clock] in the background, a [potted plant] in the background, a [vase] in the background, a [bottle] in the background, a [spoon] in the background, a [knife] in the background, a [bowl] in the background, a [banana] in the background, a [broccoli] in the background, a [wine glass] in the background, a [cup] in the background. The kitchen is well-equipped with various appliances, including a refrigerator, oven, toaster, and microwave. The refrigerator, oven, and toaster are located along the wall, while the microwave is positioned above the counter. The kitchen also features a [sink] and ample counter space, making it a functional and inviting space for meal preparation and social gatherings.
        
    \end{tabular}

\end{tabular}
\end{tcolorbox}
\vspace{-2mm}
\caption{Example captions generated with HallE-Control$_{13B}$.}
\label{tab:vis2}
\end{minipage}
\end{table*}

\begin{table*}[!h]\centering
\begin{minipage}{1.0\columnwidth}\vspace{0mm}    \centering
\begin{tcolorbox} 
\centering
\footnotesize
\begin{tabular}{p{0.97\columnwidth} c}
\ArgSty{\bf Third Case} 

\\

\begin{tabular}{ p{6cm}  p{6cm}}
      \ArgSty{\bf Image:} & \ArgSty{\bf Caption with $\varepsilon = -1$:} \\ 
     \raisebox{-\totalheight}{\includegraphics[width=0.33\textwidth]{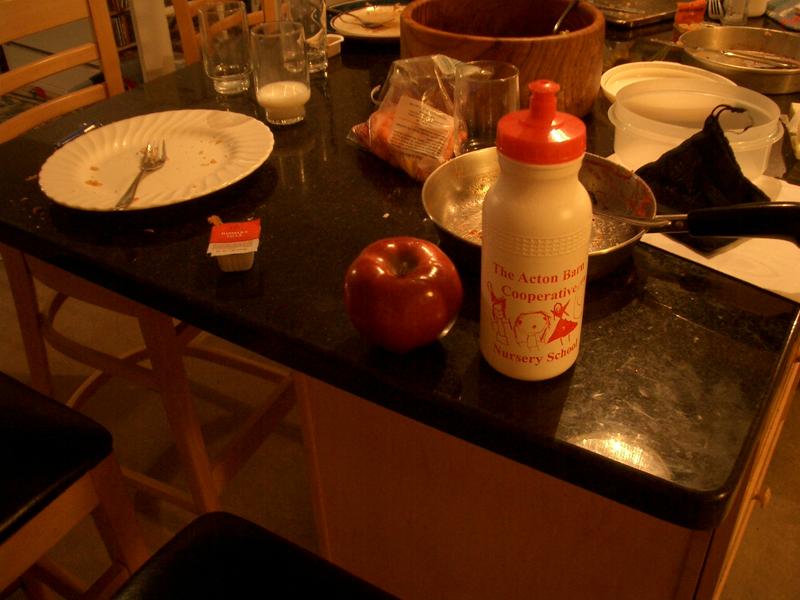}}
      & 
      The image features a kitchen counter with a few items on it. There is a bottle, an apple, and a bowl placed on the counter. The apple is situated near the middle of the counter, while the bottle is on the left side and the bowl is on the right side. In the background, there are two chairs, one near the bottom left corner and the other near the top right corner of the image. Additionally, two cups can be seen in the scene, with one placed near the bottom right corner and the other near the top center. A dining table is also visible, occupying the majority of the background.\\
      \\
      & \ArgSty{\bf Caption with $\varepsilon = 1$:} \\ 
      & The scene is set in a kitchen, where a granite countertop is cluttered with various items. There are two cups on the counter, one located towards the left side and the other towards the right. A bottle is also present on the counter, positioned towards the left side of the scene. A bowl is placed near the center of the countertop, slightly to the left. Interestingly, there is an apple and a fork on the counter as well, located towards the right side of the scene. Additionally, a stool is positioned in the background, slightly to the right of the countertop. The kitchen is equipped with two chairs, one on the far left and the other on the far right, suggesting a space for dining or socializing.
        
    \end{tabular}

\end{tabular}
\end{tcolorbox}
\vspace{-2mm}
\caption{Example captions generated with HallE-Control$_{13B}$.}
\label{tab:vis3}
\end{minipage}
\end{table*}

\begin{table*}[!h]\centering
\begin{minipage}{1.0\columnwidth}\vspace{0mm}    \centering
\begin{tcolorbox} 
\centering
\footnotesize
\begin{tabular}{p{0.97\columnwidth} c}
\ArgSty{\bf Fourth Case} 

\\

\begin{tabular}{ p{6cm}  p{6cm}}
      \ArgSty{\bf Image:} & \ArgSty{\bf Caption with $\varepsilon = -1$:} \\ 
     \raisebox{-\totalheight}{\includegraphics[width=0.33\textwidth]{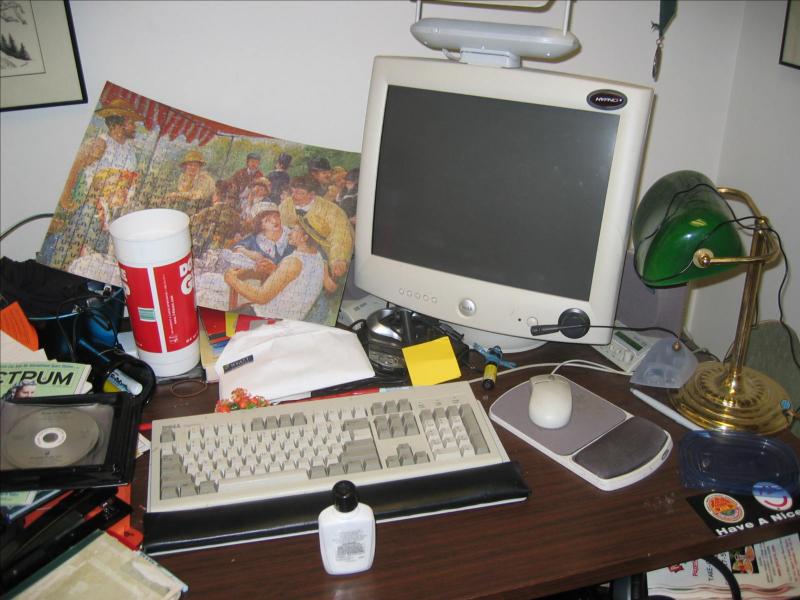}}
      & 
      The image features a computer desk with a desktop computer monitor placed on it. The monitor is surrounded by various items, including a keyboard positioned towards the front of the desk, a computer mouse located on the right side, and a [bottle] close to the monitor. There are also several cups placed on the desk, with one in the left corner, another in the middle, and the third one situated near the right side. Additionally, a pair of scissors can be found on the desk, lying horizontally between the monitor and the keyboard. The desk appears to be somewhat disorganized, giving the impression of a busy and active workspace.\\
      \\
      & \ArgSty{\bf Caption with $\varepsilon = 1$:} \\ 
      & The scene is a cluttered desk with a desktop computer setup. The computer monitor is centrally located on the desk, with a keyboard placed directly in front of it. A mouse can be found to the right of the keyboard. Various objects are scattered across the desk, including a cup on the far left, a [spoon] slightly to the right of the cup, and a [potted plant] positioned towards the left side of the desk. There are also several books spread out on the desk, with one located towards the left, another towards the right, and two more towards the center. A [handbag] is placed on the desk, located towards the right side of the scene. Despite the clutter, the desk still has a functional workspace with all necessary computer peripherals and personal items.
        
    \end{tabular}

\end{tabular}
\end{tcolorbox}
\vspace{-2mm}
\caption{Example captions generated with HallE-Control$_{13B}$.}
\label{tab:vis4}
\end{minipage}
\end{table*}

\begin{table*}[!h]\centering
\begin{minipage}{1.0\columnwidth}\vspace{0mm}    \centering
\begin{tcolorbox} 
\centering
\footnotesize
\begin{tabular}{p{0.97\columnwidth} c}
\ArgSty{\bf Fifth Case} 

\\

\begin{tabular}{ p{6cm}  p{6cm}}
      \ArgSty{\bf Image:} & \ArgSty{\bf Caption with $\varepsilon = -1$:} \\ 
     \raisebox{-\totalheight}{\includegraphics[width=0.33\textwidth]{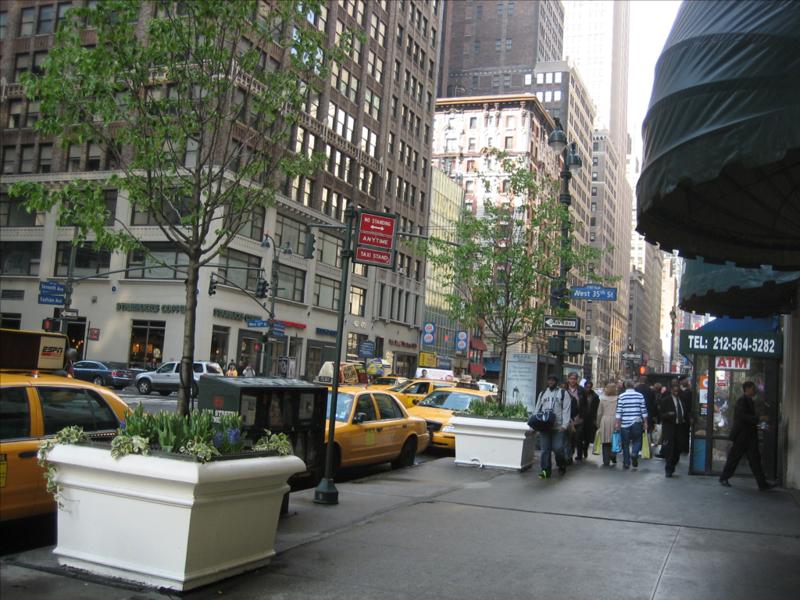}}
      & 
      "The image depicts a bustling city street with people walking down the sidewalk and various vehicles on the road. There are multiple cars, including a prominent yellow taxi cab, driving along the street. Additionally, there is a bus and a truck contributing to the busy traffic. Many pedestrians are present, some carrying handbags and backpacks, walking both individually and in groups. The scene also features a few potted plants placed along the sidewalk, adding some greenery to the urban environment. Traffic lights are visible, controlling the flow of vehicles and ensuring the safety of pedestrians.\\
      \\
      & \ArgSty{\bf Caption with $\varepsilon = 1$:} \\ 
      & The scene unfolds on a bustling city street, teeming with people and vehicles. There are numerous individuals walking along the sidewalk, some closer to the foreground, while others are further away. The street is filled with a variety of vehicles, including cars, [trucks], and taxis. One car is positioned towards the left side of the scene, while another is located more towards the center. Two [trucks] are also present, one near the middle of the street and the other towards the right side. Taxis are a common sight on this busy street, with one taxi near the middle of the scene and another towards the right side of the street. The cityscape is further enhanced by the presence of a fire hydrant on the sidewalk, adding to the urban atmosphere. The scene captures the essence of a typical day in a bustling city, with people and vehicles navigating their way through the streets.
        
    \end{tabular}

\end{tabular}
\end{tcolorbox}
\vspace{-2mm}
\caption{Example captions generated with HallE-Control$_{13B}$.}
\label{tab:vis5}
\end{minipage}
\end{table*}

\subsection{Theoretical Explanations}
\textbf{HallE-Control Formulation}

HallE-Control follows LLaVA's training strategy, which freezes vision encoder and language model, only finetune the projector for the first stage.
The visual encoder $g(\cdot)$ transfer image $X_v$ to visual features: $$Z_v = g(X_v)$$
The projector $W$ connects image feature to the word embedding space: $$H_v = W \cdot Z_v$$.

In the second stage, both projector and language model is trained. The model is trained on two type of data: 1. Contextual only data. 2. Contextual and parametric combined data.

Language model trained on contextual only data has a distribution initiating from $\pi$. The other language model trained on combined data has a distribution initiating from $\pi \textquotesingle$ 

With the Theorem 1 in LM-Switch~\citep{han2023lm}, 
under the same assumptions, there exists an matrix $W$, transforming a word embedding $E$ to $WE$, which is equivalent to let a LM simulate the text distribution initiating from another distribution.

Inspired by the Theorem, we propose a linear transform in word embedding space for LMM. Let $M$ be the finetuned LLM, with control layer/projector denoted as $W$, we replace each word embedding $e_v$ with $e_v + \varepsilon We_v$, making the new language model $M' = M(\varepsilon W)$ as HallE-Control's language decoder. $\varepsilon$ is adjustable from -1 to 1. We assign $\varepsilon=-1$ and fit contextual only data; assign $\varepsilon=1$ with the contextual and parametric combined data. After finetuning the HallE-Control, the user only needs to specify a control value $\varepsilon \in [-1, 1]$ and do normal vision language task like image captions. We use maximal likelihood as the training objective.

\textbf{Continuous Control}

The design of LM-Switch maintains a linearity guarantee, with proof of the control model's distribution is close to a linear interpolation.
Let $\lambda_{max}$ be the maximum eigen-value of W. When varying $\varepsilon \textquotesingle$s value,

$$
\lVert P(\cdot|k\varepsilon, W) - (P(\cdot)(1-k) + kP(\cdot |\varepsilon, W))\rVert_1 \le 2\lvert k(1-k)\rvert \varepsilon^2L^2\lambda_{max}(e^{\lambda_{max}}-1)
$$
distribution of the control model is close a linear interpolation of $M$ and $M'$, meaning that the model distribution changes linearly. Therefore, our method can take any $\varepsilon$ between 1 and -1.

For HallE-Control, the core idea is: Assuming LLM is good enough to represent an equivalent distribution with HMM; there exist an matrix W, so that after transferring word embedding $E$ to $WE$, the LLM's originally simulate the text distribution starting with initial state $\pi$ will turn to be equivalent to a distribution starting with initial state $\pi^{\prime}$.

According to experiments in Section 4.1, it reveals language models can distinguish inference object and observed objects by adding special tokens. Because, decoder based language models generate sequences in an auto-regressive way. We followed the proof in \cite{han2023lm}:
if we assume \textbf{O} as our observation space which means a set of m observations. We assume $\textbf{c}(o_1, ..., o_{t-1})$ as a contextual vector, $\textbf{E}=(e_o, ...) \in \mathbb{R^{d\times|\mathcal{O}|}} $ as word embedding. We can represent word logits as $\textbf{l} = \textbf{c}(o_1, ..., o_{t-1})^\top$\textbf{E}. In attention mechanic, it will pass through a softmax operator to get the distribution over words, We use the same assumption in LM-Switch method to assume a linear formulation and let the conditional probability in language model $P(o_t|o_1, ...,o_{t-1}) = \textbf{c}(o_1, ..., o_{t-1})^\top e_{o_t}$. We can get the full probability by using chain-rule: $\prod_{t=1}^{T}P(o_t|o1, ...,o_{t-1}) = P(0_1, ...o_{t-1})$. We assume our LLMs are good enough to represent an equivalent distribution with HMM and full column-rank for \textbf{E}, \textbf{p}(o).

Our conclusion is: Applying a linear transformation on word embedding space is equivalent to a shift from one initial condition to another. This is the reason that we want to shift a language model with distribution produce higher inference imagination conditional probability to a lower one. 



\end{document}